\documentclass[11pt,letter]{article}

\newif\ifPS
\PStrue
\PSfalse

\usepackage{epsfig}
\usepackage{comment}
\usepackage{times}
\usepackage{helvet}
\usepackage{courier}
\usepackage{psfrag}
\usepackage{fleqn}
\usepackage{latexsym}
\usepackage{amsmath}
\usepackage{amsfonts}
\usepackage{amssymb}
\usepackage[T1]{fontenc}
\usepackage{plain}
\usepackage{MnSymbol}
\usepackage[normal]{threeparttable}
\usepackage{cclicenses}

\usepackage{tikz}
\usetikzlibrary{positioning,arrows}

\definecolor{g}{gray}{0.85}

\newtheorem{definition}{Definition}
\newtheorem{theorem}{Theorem}

\newcommand{\plan}{{\pi}}

\newcommand{\Act}{{\cal A}}

\newcommand{\R}{I\hspace{-1.3mm}R}

\newcommand{\N}{I\hspace{-1.3mm}N}
\newcommand{\pset}{\powerset_{\hspace{-1mm} s}}
\newcommand{\EG}{{\cal E} }
\newcommand{\X}{{\cal X}}
\newcommand{\K}{{\cal K}}
\newcommand{\Op}{{\cal O}p}
\newcommand{\nOp}{{\cal O}p{\hspace{-1.5mm}}{'}~}
\newcommand{\Prop}{{\cal P}_{\hspace{-0.7mm} r}}

\newcommand{\case}[1]{\langle \Pi_{#1}, \pi_{#1} \rangle}

\newcommand{\lpg}{\mbox{\sc lpg}}
\newcommand{\OAKplan}{\mbox{\sc OAKplan}}
\newcommand{\OBJMATCH}{\mbox{\tt obj\_match}}

\newcommand{\sgplan}{\mbox{\sc sgplan6}}

\newcommand{\PEG}{\EG}

\newcommand{\citeyear}{\cite}

\newif\ifTR
\TRtrue

\begin{document}

\date{}

\title{Case-Based Merging Techniques in \OAKplan }

\author{
FUB
\ifTR
\\
\\
{\bf Technical Report }
\fi
}

\maketitle

\thispagestyle{empty}

\begin{abstract}

Case-based planning can take advantage of former problem-solving
experiences by storing in a plan library previously generated plans
that can be reused to solve similar planning problems in the future.
Although comparative worst-case complexity analyses of plan
generation and reuse techniques reveal that it is not possible to
achieve provable efficiency gain of reuse over generation, we show
that the case-based planning approach can be an effective
alternative to plan generation when {\em similar} reuse candidates
can be chosen.
 

\vspace{1.8cm}
\hspace{-0.8cm}
{
\small
\noindent
{\em Keywords}:
Case-Based Planning, Domain-Independent Planning, Merging Techniques, Case-Based Reasoning, Heuristic Search for Planning, Kernel Functions.
}

\end{abstract}

\ifTR
{\Huge
\begin{center}
\cc
\byncnd
\end{center}
}

\newpage
\tableofcontents
\vspace{1.8cm}
\newpage

\fi

 \section{Introduction}
This report describes a case-based planning system called
\textsc{OAKplan}, with a special emphasis on newly implemented merging
techniques. This extension was motivated by the observation that,
especially in some domains, the performance of the system can be
greatly improved by remembering some \emph{elementary} solutions for a
simple problems, which can be combined to address significantly more
complex scenarios.

 \section{Preliminaries}
\label{sec:Preliminaries}
 In the following we present some notations
 that will be used in the paper with an analysis of
 the computational complexity of the problems considered.

 \subsection{Planning formalism}

Similarly to Bylander's work \cite{Byl94}, we define an instance of
propositional planning problem as:

\begin{definition} An instance of {\bf propositional STRIPS planning problem}
is a tuple $\Pi =\langle \Prop,{\cal I, G}, \Op \rangle$ where:
\begin{itemize}
\item ${\Prop}$ is a finite set of ground atomic propositional formulae;

\item ${\cal I} \subseteq{\Prop}$ is the initial state;

\item ${\cal G} \subseteq{\Prop}$ is the goal
specification;

\item $\Op$ is a finite set of operators, where each operator $o \in \Op$ has the form $o^p \Rightarrow o_+, o_-$ such that
\begin{itemize}
\item $o^p \subseteq{\Prop}$ are the propositional preconditions,
\item $o_+ \subseteq{\Prop}$ are the positive postconditions (add list),
\item $o_- \subseteq{\Prop} $ are the negative postconditions (delete list)
 \end{itemize}
and $o_+ \cap o_- = \emptyset $.

\end{itemize}
\end{definition}

We assume that the set of propositions ${\Prop}$ has a particular
structure. Let {\bf O} be a set of {\em typed} constants $c_i$, with
the understanding that distinct constants denote distinct objects
(corresponding to {\em individual entities} following the Conceptual
Graphs notation \cite{Chein2008}). Let ${\bf P}$ be a set of
predicate symbols, then ${\Prop}({\bf O, P})$ is the set of all
ground atomic formulae over this signature. Note that we use a {\it
  many-sorted logic} formalisation since it can significantly increase
the efficiency of a deductive inference system by eliminating useless
branches of the search space of a domain
\cite{Chien98,Cohn87,Walther85}. A fact is an assertion that some
individual entities exist and that these entities are related by some
relationships.

A plan $\pi$ is a partially ordered sequence of actions $\pi=(a_1, \dots,a_m,
{\cal C})$, where $a_i$ is an action (completely instantiated operator) of
$\pi$ and ${\cal C}$ defines the ordering relations between the actions of
$\pi$.
A {\it linearisation} of a partially ordered plan is a total order
over the actions of the plan that is consistent with the existing
partial order.  In a totally ordered plan $\pi=(a'_1, \dots,a'_m)$, a
precondition $f$ of a plan action $a'_i$ is {\it supported} if {\em (i)} $f$
is added by an earlier action $a'_j$ and not deleted by an intervening
action $a'_k$ with $j< k<i$ or {\em (ii)} $f$ is true in the initial
state and $\not \exists a'_k$ with $k<i$ s.t. $f\in del(a'_k)$. In a
partially ordered plan, a precondition of an action is {\it possibly
  supported} if there exists a linearisation in which it is supported, while an
action precondition is {\it necessarily supported} if it is supported in all
linearisations. An action precondition is {\it necessarily unsupported} if
it is not possibly supported. A {\it valid plan} is a plan in which all
action preconditions are necessarily supported. \\ \\
The complexity of STRIPS planning problems has been studied extensively in the
literature. Bylander \citeyear{Byl94} has defined PLANSAT as the decision problem
of determining whether an instance $\Pi$ of propositional STRIPS planning has
a solution or not. PLANMIN is defined as the problem of determining if there
exists a solution of length $k$ or less, i.e., it is the decision
problem corresponding to the problem of generating plans with minimal
length. Based on this framework, he has analysed the computational complexity of a
general propositional planning problem and a number of generalisations and
restricted problems. In its most general form, both PLANSAT and PLANMIN are
PSPACE-complete. Severe restrictions on the form of the operators are
necessary to guarantee polynomial time or even NP-completeness \cite{Byl94}.\\ \\
To address the high complexity of the planning problems, different heuristical approaches arise. The case based approach relies on encountering problems similar to each other and tries to reuse previously found plans to solve new problems. If successful, this can save a considerable amount of resources. Clearly, to apply such a technique, we need similar problems to have similar solution --- in other words, we need the world to be \emph{regular}. When solving a new problem, the planner searches previously solved problems and retrieves the most suitable one which is then adapted to solve the current problem.

In general, solved problems are stored in a case base, which is a collection of cases; a case is a pair consisting of a problem description and its solution. Following the formalisation of Liberatore \cite{Liberatore05}, we define a case base as follows:
\begin{definition} A \emph{case base}, or a plan library, is a set $\lbrace \case{i}, 1 \leq i \leq n \rbrace$ where each $\case{i}$ is a \emph{planning case} with $\Pi_i$ being an instance of propositional STRIPS planning problem and $\pi_i$ a solution to $\Pi_i$. 
\end{definition}
Note that different planners implement cases differently --- several solution plans may be stored instead of just one, justifications may be added, some planners even avoid storing a plan as a set of actions and store its derivational trace instead. 

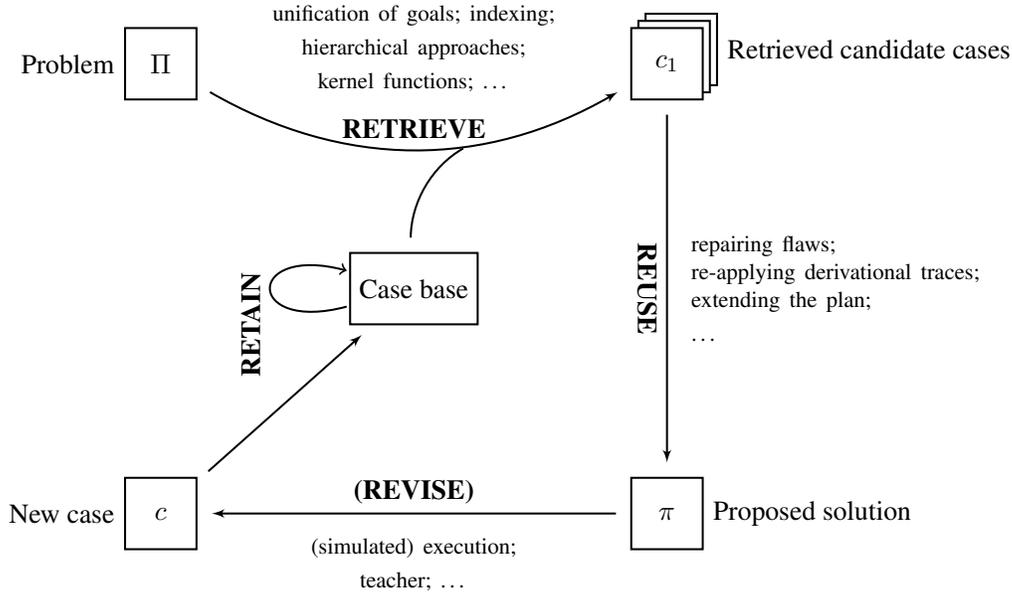
\begin{figure}[h!]
   \centering
   \scalebox{.95}{
   \begin{tikzpicture}[case/.style={draw,thick,fill=white,minimum size=1cm}, %
     node distance=3cm, conn/.style={-latex', shorten >=2mm, shorten <=2mm, thick}]
     \node [case] (cb) at (0,0) {Case base};
     \node [case, above left=of cb, label={left:Problem}] (Pi) {$\Pi$};
     \node [case, above right=of cb, xshift=2mm, yshift=2mm, label={right:Retrieved candidate cases}] {};
     \node [case, above right=of cb, xshift=1mm, yshift=1mm] {};
     \node [case, above right=of cb] (c1) {$c_1$};
     \node [case, below right=of cb, label={right:Proposed solution}] (pi) {$\plan$};
     \node [case, below left=of cb, label={left:New case}] (c) {$c$};

     \path (Pi) edge [conn, bend right] node [above,midway] (retrieve) {\textbf{RETRIEVE}} (c1);
     \node [above=-4mm  of retrieve, xshift=10mm, yshift=-1mm] (X) {};
     \path (cb) edge [conn, -, bend left] (X);
     \node [above=1mm of retrieve, text width=13em, align=center] {\footnotesize unification of goals; indexing;\\hierarchical approaches; kernel functions; \textellipsis};

     \path (c1) edge [conn] node [below,midway,sloped] {\textbf{REUSE}} %
     node [right,midway,text width=14em,xshift=2mm] {\footnotesize repairing flaws;\\re-applying derivational traces;\\extending the plan;\\\textellipsis} (pi);

     \path (pi) edge [conn] node [above] {\textbf{(REVISE)}} %
     node [below,text width=13em, align=center, yshift=-2mm] {\footnotesize (simulated) execution;\\teacher; \textellipsis} (c);

     \path (c) edge [conn] (cb);
     \path (cb) edge [loop left, conn, shorten <=1pt] node [above, sloped, xshift=-5mm] {\textbf{RETAIN}} (cb);

     \path [use as bounding box] (-6.3,-4.1) rectangle (7.2,4.2);
   \end{tikzpicture}
 }
   \caption{\small{ \textbf{The case-based cycle.} %
	The system queries the case base to retrieve case(s) useful for solving the problem $\Pi$ (\textbf{retrieve}); the solution(s) from the retrieved case(s) are applied to $\Pi$ in order to find its solution $\plan$ (\textbf{reuse}); the new solution is revised to identify and correct possible flaws (\textbf{revise}); the verified solution $\plan$ and the problem $\Pi$ are used to create a new case that may be introduced to the case base, in addition the system may be capable of auto-analysis of the case base when reaching some limit conditions (\textbf{retain}).}
   }
   \label{fig:cb-cycle}
\end{figure}

In order to realise the benefits of remembering and reusing past
solutions, a case-based system needs to efficiently implement several
procedures, such as those for retrieving analogous cases
(\emph{Retrieve}), adapting them to suit the new problem (\emph{Reuse,
  Revise}), and building and maintaining a case base of sufficient
size and coverage to yield useful analogues (\emph{Retain}). When a
case-based system faces a new problem, it performs these procedures in
a sequence that starts by querying the case base and ends by
(possibly) updating it (Fig.~\ref{fig:cb-cycle}).

Regardless of the underlying formalisation, there are two main approaches to case-based planning, whose fundamental difference is in the way they adapt the cases to solve the current problem. A \emph{conservative} or a  \emph{transformational} plan adaptation inisists on reusing as much of the starting plan $\pi_i$ as possible. It turns out that such adaptation may be very expensive. Moreover, the quality of the solution strongly depends on the correspondence between the retrieved case and the current problem, which is influenced by the way the case is retrieved from the case base as well as by the case base itself, or rather by its competence. Intuitively, a \emph{competence} is a feature of the case base referring to the variety of problems for which the case base contains a reasonably good case to adapt to a solution; the more problems the case base can address, the higher competence it has. The \emph{generative} approach, on the contrary, treats the case as a hint that can even be fully ignored and the solution can be generated from scratch. Hence the competence of the case base is not so crucial for the quality of the solution plan. On the other hand, such process may be worse than traditional plan-generation in terms of complexity, as it first needs to search the case base to possibly ignore what has been found and to generate a whole new plan. This shows that the competence of the case base is also important here, though it influences the complexity of finding the solution rather than its quality. 

It is important to remark that our approach is more related to the
generative case-based approach than to the transformational, in that the plan library is only assumed to be used when it
is useful in the process of searching for a new plan, and is not
necessarily used to provide a starting point of the search process. If
a planning case $\case{*}$ is also known, the
current planning problem $\Pi$ cannot become more difficult, as we can simply disregard the
case and find the plan using $\Pi$ only.


Essential to this trivial
result is that, similarly to most modern plan adaptation and
case-based planning approaches
\cite{DanaNau2002,GerSer-aips00,TonidandelR02,KrogtW-icaps05}, we do
not enforce plan adaptation to be conservative, in the sense that we
do not require to reuse as much of the starting plan $ \pi_*$ to solve
the new plan.  The computational complexity of plan Reuse and
Modification for STRIPS planning problems has been analysed in a
number of papers
\cite{Byl94,Bylander96,KuchibatlaM06,Liberatore05,Nebel95}.

Moreover empirical analyses show that plan modification for similar planning
instances is somewhat more efficient than plan generation in the average
case
\cite{Byl93,fox-icaps06,GerSer-aiia99,GerSer-aips00,Nebel95,Srivastava07,KrogtW-icaps05}. \\ \\
\noindent It is crucial that the system has at its disposal a competent case base.
Intuitivelly, the \emph{competence} of the case
base measures how often the case base provides a good reuse
candidate. Of course, the number of successful reuses also depends on
other components of ths system (how exactly is the case base queried,
what is the reuse strategy, when do we consider the solution a
``successful reuse'', etc.). 

The competence
of the case base grows as we keep adding solutions to new diverse
problems, but only to a certain point when the query over the case
base takes too much time to be answered. If the system works in the
environment where the problems tend to be very complex, it is
unfortunately very unlikely that the case base could contain
sufficient number of cases to cover majority or at least a significant
part of the kinds of problems to be solved. One of the promising
directions for such scenarios is to attempt to build a case base of
``elementary'' cases which address only simple sub-problems and which
are combined together to address also complex problems. The number of
such elementary problems is lower and hence it is easier to achieve a
competent case base.

Clearly such modification needs to be accompanied by the change of the retrieval policy --- we no longer look for the most similar case, but for a suitable set of partial solutions. The partial solutions are then put together; depending on the interaction between the subgoals a simple concatenation may be sufficient, but in case the subgoals interfere the concatenation may be very inefficient (e.g. in the logistics domain).

\subsection{Graphs}

Graphs provide a rich means for modelling structured objects and they
are widely used in real-life applications to represent molecules,
images, or networks.  On a very basic level, a graph can be defined by
a set of entities and a set of connections between these entities.
\ifTR
Due to their universal definition, graphs have found widespread
application for many years as tools for the representation of complex
relations. Furthermore, it is often useful to compare objects
represented as graphs to determine the degree of similarity between
the objects.  
\fi
More formally:

\begin{definition}
A {\bf labeled graph} $G$ is a 3-tuple $G=(V,E, \lambda)$, in which
\begin{itemize}
\item $V$ is the set of vertices,
\item $E\subseteq V\times V$ is the set of directed edges or arcs,
\item $\lambda:V\cup E \rightarrow \pset(L_\lambda)$ is a function assigning
 labels to vertices and edges;
\end{itemize}
\end{definition}
where $L_\lambda$ is a finite set of symbolic labels and
$\pset(L_\lambda)$ represents the set of all the {\it multisets} on
$L_\lambda$. Note that our label function considers {\it multisets} of
symbolic labels, with the corresponding operations of union,
intersection and join \cite{Blizard89}, since in our context they are
more suitable than standard sets of symbolic labels in order to
compare vertices or edges accurately as described later.  The above
definition corresponds to the case of directed graphs; undirected
graphs are obtained if we require for each edge $[v_1, v_2]\in E$ the
existence of an edge $[v_2, v_1]\in E$ with the same label.
$|G|=|V|+|E|$ denotes the {\it size} of the graph $G$, while an empty graph
such that $|G|=0$ will be denoted by $\emptyset$.
An arc $e = [v,u]\in E$ is considered to be directed from $v$ to $u$;
$v$ is called the {\it source} node and $u$ is called the {\it target}
node of the arc; $u$ is said to be a {\it direct successor} of $v$,
$v$ is said to be a {\it direct predecessor} of $u$, while $v$ is said
to be {\it adjacent} to the vertex $u$ and vice versa.

Here we present the notion of graph union which is essential for the
definition of the graphs used by our matching functions:
\begin{definition}
  The {\bf union} of two graphs $G_1=(V_1,
  E_1,\lambda_1)$ and $G_2=(V_2, E_2,\lambda_2)$, denoted by $G_1 \cup G_2$, is the graph $G=(V, E,\lambda)$ defined
  by
 \begin{itemize}
 \item $V=V_1 \cup V_2  $,
 \item $E=E_1 \cup E_2 $,
 \item
 $\displaystyle \lambda(x) =\left\{
 \begin{array}{ll}
 
 \lambda_1(x) & {\tt if~} x \in (V_1 \backslash
 V_2) \vee x \in (E_1 \backslash E_2) \\
 \lambda_2(x) &  {\tt if~} x \in (V_2 \backslash
 V_1) \vee x \in (E_2 \backslash E_1) \\
\lambda_1(x) \uplus \lambda_2(x) \hspace{10mm} & {\tt otherwise}
 \end{array}
 \right.
 $
\end{itemize}
\end{definition}

where $\uplus$ indicates the join, sometimes called {\it sum}, of two
multisets \cite{Blizard89}, while $\lambda(\cdot)$ associates a multiset of
symbolic labels to a vertex or to an edge.

In many applications it is necessary to compare objects represented as
graphs and determine the similarity among these objects.  This is
often accomplished by using graph matching, or isomorphism techniques.
Graph isomorphism can be formulated as the problem of identifying a
one-to-one correspondence between the vertices of two graphs such that
an edge only exists between two vertices in one graph if an edge
exists between the two corresponding vertices in the other graph.
Graph matching can be formulated as the problem involving the maximum common
subgraph (MCS) between the collection of graphs being considered. This is
often referred to as the maximum common substructure problem and denotes the
largest substructure common to the collection of graphs under consideration.
More precisely:
\begin{definition}
 Two labeled graphs $G = (V, E, \lambda)$ and $G'= (V', E', \lambda')$ are
 {\bf isomorphic} if there exists a bijective function $f : V
 \rightarrow V'$ such that
 \begin{itemize}
 \item $\forall v\in V, \lambda(v)=\lambda' \bigl(f(v)\bigr)$,
 \item $ \forall [v_1, v_2]\in E, \lambda \bigl([v_1,
 v_2]\bigr)=\lambda'\bigl( \left[f(v_1),
 f(v_2)\right] \bigr)$,
 \item $[u, v]\in E$ if and only if ~ $\bigl[f (u), f (v)\bigr]\in
 E' $.
 \end{itemize}
 We shall say that $f$ is an isomorphism function.
\end{definition}
\begin{definition} An {\bf Induced Subgraph} of a graph $G= (V, E, \lambda)$ is a graph
 $S=(V', E', \lambda')$ such that
\begin{itemize}
\item $V'\subseteq V$ and $\forall v\in V'$,   $\lambda'(v)\subseteq \lambda(v)$,
\item $E' \subseteq E $ and $\forall e \in E'$,  $\lambda'(e)\subseteq \lambda(e)$
\item $\forall v,u \in V'$, $[v,u]\in E'$ if and only if $[v,u]\in E $
\end{itemize}
\end{definition}

A graph $G$ is a {\it Common Induced Subgraph} ({\it CIS}) of
graphs $G_1$ and $G_2$ if $G$ is isomorphic to induced subgraphs
of $G_1$ and $G_2$.
A common induced subgraph $G= (V, E, \lambda)$ of $G_1$ and $G_2$ is
called {\it Maximum Common Induced Subgraph} ({\it MCIS}) if there
exists no other common induced subgraph of $G_1$ and $G_2$ with
$\sum_{v\in V } | \lambda(v)| $ greater than $G$.
Similarly, a common induced subgraph $G= (V, E, \lambda)$ of $G_1$ and
$G_2$ is called {\it Maximum Common Edge Subgraph} ({\it MCES}), if
there exists no other common induced subgraph of $G_1$ and $G_2$ with
$\sum_{e\in E } | \lambda(e)| $ greater than $G$.
Note that, since we are considering multiset labeled
graphs, we require a stronger condition than standard MCIS and MCES
for labeled graph, in fact we want to maximise the total cardinality of the
multiset labels of vertices/edges involved instead of the simple
number of vertices/edges.

As it is well known, subgraph isomorphism and MCS between two or among
more graphs are NP-complete problems \cite{GareyJohnson79}, while it
is still an open question if also graph isomorphism is an NP-complete
problem.  As a consequence, worst-case time requirements of matching
algorithms increase exponentially with the size of the input graphs,
restricting the applicability of many graph based techniques to very
small graphs.

 \subsection{Kernel Functions for Labeled Graphs}

 In recent years, a large number of graph matching methods based on different
 matching para\-digms have been proposed, ranging from the spectral
 decomposition of graph matrices to the training of artificial neural networks
 and from continuous optimisation algorithms to optimal tree search
 procedures.

 The basic limitation of graph matching is due to the lack of any
 mathematical structure in the space of graphs. Kernel machines, a new
 class of algorithms for pattern analysis and classification, offer an
 elegant solution to this problem \cite{Scholkopf2001}.  The basic
 idea of kernel machines is to address a pattern recognition problem
 in a related vector space instead of the original pattern space. That
 is, rather than defining mathematical operations in the space of
 graphs, all graphs are mapped into a vector space where these
 operations are readily available. Obviously, the difficulty is to
 find a mapping that preserves the structural similarity of graphs, at
 least to a certain extent. In other words, if two graphs are
 structurally similar, the two vectors representing these graphs
 should be similar as well, since the objective is to obtain a vector
 space embedding that preserves the characteristics of the original
 space of graphs.

 A key result from the theory underlying kernel machines states that
 an explicit mapping from the pattern space into a vector space is not
 required. Instead, from the definition of a kernel function it
 follows that there exists such a vector space embedding and that the
 kernel function can be used to extract the information from vectors
 that is relevant for recognition.  As a matter of fact, the family of
 kernel machines consists of all the algorithms that can be formulated
 in terms of such a kernel function, including standard methods for
 pattern analysis and classification such as principal component
 analysis and nearest-neighbour classification. Hence, from the
 definition of a graph similarity measure, we obtain an implicit
 embedding of the entire space of graphs into a vector space.

A {\it kernel function} can be thought of as a special similarity measure
with well defined mathematical properties \cite{Scholkopf2001}.
Considering the graph formalism, it is possible to define a kernel
function which measures the degree of similarity between two
graphs. Each structure could be represented by means of its similarity
to all the other structures in the graph space.  Moreover a kernel
function implicitly defines a {\em dot product} in some space
\cite{Scholkopf2001}; i.e., by defining a kernel function between two
graphs we implicitly define a vector representation of them without
the need to explicitly know about it.

From a technical point of view a kernel function is a special similarity measure $k: \X
\times \X \rightarrow \R$ between patterns lying in some arbitrary
domain $\X$, which represents a dot product, denoted by $ \langle
\cdot, \cdot \rangle$, in some Hilbert space ${\cal H}$
\cite{Scholkopf2001}; thus, for two arbitrary patterns $x, x' \in \X$
it holds that $k(x, x')=\langle \phi(x),\phi(x')\rangle$, where $\phi:
\X \rightarrow {\cal H}$ is an arbitrary mapping of patterns from the
domain $\X$ into the feature space ${\cal H}$. In principle the
patterns in domain $\X$ do not necessarily have to be vectors; they
could be strings, graphs, trees, text documents or other objects. The
vector representation of these objects is then given by the map
$\phi$.  
Instead of performing the expensive transformation step explicitly,
the kernel can be calculated directly, thus performing the feature
transformation only implicitly: this is known as  {\it kernel
  trick}.  This means that any set, whether a linear space or not,
that admits a positive definite kernel can be embedded into a linear
space.

 \begin{figure*}[tbp]
 \begin{center}
 \includegraphics[angle=0,width=0.9\textwidth]{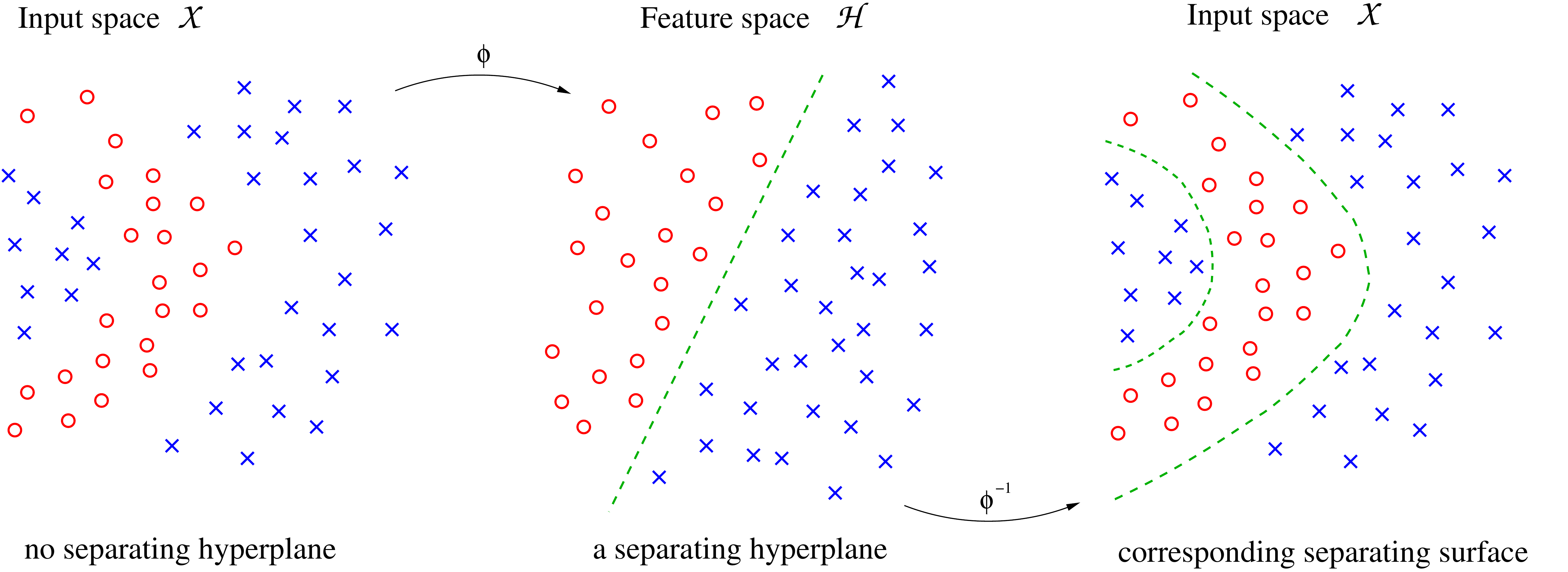}
 \end{center}
 \vspace{-6mm}
 \caption{\ifTR \small \fi \label{fig:separating-hyperplan}
The kernel approach for classification. Left: non-linearly separable
input provided by dots and crosses. Middle: perfect or approximate
linear-separability can be achieved in feature space via the mapping
$\phi$. Right: linear decision surface in feature space defines a complex
decision surface in input space.
}
 \vspace{-2mm}

 \end{figure*}

More specifically, kernel methods manage non-linear complex tasks making use of
linear methods in a new space. For instance, take into consideration a
classification problem with a training set ${\cal S} = \{(u_1 , y_1 ),
\dots ,$ $(u_n , y_n )\}$, $(u_i , y_i) \in {\cal X} \times Y$, for $i
= 1, \dots , n$, where ${\cal X}$ is an inner-product space  $ \Large\left( \normalsize\textrm{i.e. } \R^d\Large\right)$
 and $Y =
\{ -1, +1\}$. In this case, the learning phase corresponds to building a
function $f \in Y^{\cal X}$ from the training set ${\cal S}$ by
associating a class $y \in Y$ to a pattern $u \in {\cal X}$ so that
the generalisation error of $f$ is as low as possible.

A functional form for $f$ consists in the hyperplane $f (u) = sign
(\langle w, u \rangle + b)$, where $sign (\cdot)$ refers to the function
returning the sign of its argument. The decision function $f$ produces
a prediction that depends on which side of the hyperplane $\langle w, u
\rangle + b = 0$ the input pattern $u$ lies.
The individuation of the {\em best} hyperplane corresponds to a convex
quadratic optimisation problem in which the solution vector $w$ is a
linear combination of the training vectors:

\noindent
\hspace{3.5cm}$w =\sum_{i=1}^n \alpha_i y_i u_i$, for some $\alpha_i \in \R^+ $, $i =
1, \dots ,n$.

In this way the linear classifier $f$ may be rewritten as
\[f (u) = sign
\left( \sum_{i=1}^n \alpha_i y_i \langle u_i, u \rangle + b\right)
\]

As regards  complex classification problems, the set of all possible linear
decision surfaces might not be rich enough in order to provide a good
classification, independently  from the values of the parameters $w
\in{\cal X} $ and $b \in \R$ (see Figure
\ref{fig:separating-hyperplan}). The aim of the kernel trick is
that of overcoming this limitation by adopting a linear approach to
transformed data $\phi (u_1 ), \dots , \phi(u_n )$ rather than
raw data. Here $\phi$ indicates an embedding function from the input
space ${\cal X}$ to a feature space ${\cal H}$, provided with a dot
product.
This transformation enables us to give an alternative
kernel representation of the data which is equivalent to a mapping into a
high-dimensional space where the two classes of data are more readily
separable. The mapping
is achieved through a replacement of the inner product:
\[ \langle u_i, u \rangle \rightarrow \langle \phi(u_i), \phi(u) \rangle  \]
and the separating function
can be rewritten as:
\begin{equation}\label{eq:f(u)}
f (u) = sign
\left( \sum_{i=1}^n \alpha_i y_i \langle \phi(u_i), \phi(u) \rangle + b\right)
\end{equation}

The main idea behind the kernel approach consists in replacing the dot
product in the feature space using a kernel $k(u, v) = \langle
\phi(v), \phi(u) \rangle$; the functional form of the mapping
$\phi(\cdot)$ does not actually need to be known since it is
implicitly defined by the choice of the kernel.
A positive definite kernel \cite{Gartner2003} is:
\begin{definition}
Let $\X$ be a set. A symmetric function
$k : \X  \times \X \rightarrow  \R$
is a positive definite {\bf kernel function} on $\X$ iff~~
$\forall n \in \N$,  $\forall x_1 , \dots , x_n \in \X$, and
$\forall c_1 , \dots , c_n \in \R$
\[ \sum_{i,j\in \{1,...,n\}} c_i c_j k(x_i , x_j ) \geq 0\]
\end{definition}
where $\N$ is the set of positive integers.
For a given set $S_u = \{u_1 , \dots , u_n \}$, the matrix ${\bf K}=\left( k(u_i, u_j) \right)_{i,j}$ is known as Gram
matrix of $k$ with respect to $S_u$. Positive definite kernels are
also called Mercer kernels.

\begin{theorem}[ Mercer's property \cite{Mercer1909}] \label{th:mercer}
For any positive definite kernel function
$k \in \R^{\X \times \X }$, there exists a mapping $\phi \in {\cal H}^\X$ into the feature space ${\cal H}$ equipped
with the inner product $\langle \cdot, \cdot \rangle_{\cal H}$, such that:
\[ \forall u, v \in \X, ~~~~~~~~~   k(u, v) = \langle \phi(u), \phi(v)\rangle_{\cal H}
\]
\end{theorem}

The kernel approach replaces all inner products in
Equation \ref{eq:f(u)} and all related expressions to compute the
real coefficients $\alpha_i$ and $b$, by means of a Mercer kernel $k$. For any
input pattern $u$, the relating decision function $f$ is given by:
\begin{equation}\label{eq:kf}
f (u) = sign
\left( \sum_{i=1}^n \alpha_i y_i k(u_i, u) + b\right)
\end{equation}

This approach transforms the input patterns $u_1 , \dots , u_n$ into
the corresponding vectors $\phi(u_1 ), \dots , \phi(u_n ) \in {\cal
  H}$ through the mapping $\phi \in {\cal H}^\X$ (cf. Mercer's
property, Theorem \ref{th:mercer}), and uses hyperplanes in the
feature space ${\cal H}$ for the purpose of classification (see Figure
\ref{fig:separating-hyperplan}).  The dot product $\langle u, v\rangle
= \sum^d_{i=1} u_i v_i$ of $\R^d$ is actually a Mercer kernel, while
other commonly used Mercer kernels, like polynomial and Gaussian
kernels, generally correspond to nonlinear mappings $\phi$ into
high-dimensional feature spaces ${\cal H}$. On the other hand the Gram
matrix implicitly defines the geometry of the embedding space and
permits the use of linear techniques in the feature space so as to
derive complex decision surfaces in the input space $\X$.

While it is not always easy to prove positive definiteness for a given kernel,
positive definite kernels are characterised by interesting closure properties.
More precisely, they
are closed under sum, direct sum, multiplication by a scalar, tensor
product, zero extension, pointwise limits, and exponentiation
\cite{Scholkopf2001}.
\ifTR
Well-known examples of kernel functions are:\\
\begin{tabular}{l l}
$\bullet$
 Radial Basis Functions\hspace{3cm} &
$k_{RBF}(x,x') = exp\left(\frac{-||x-x'||^2}{2\sigma^2}\right)$;\\
$\bullet$
Homogenous polynomial kernels\hspace{1cm}& $k_{poly}(x,x')=\langle x, x'\rangle^d $ ($d \in \N$);\\
$\bullet$
Sigmoidal kernels\hspace{1cm}& $k_{Sig}(x,x') =tanh \left(\kappa (x \cdot x')+ \theta \right)$\\
$\bullet$
Inv. multiquadratic  kernels\hspace{1cm}& $k_{inv}(x,x') =\frac{1}{\sqrt{||x-x'||^2 + c^2}}   $
\end{tabular}
\fi

A remarkable contribution to graph kernels is the work on convolution
kernels, that provides a general framework to deal with complex
objects  consisting of simpler parts
\cite{haussler99convolution}. Convolution kernels derive the similarity
of complex objects from the similarity of their parts.  Given two
kernels $k_1$ and $k_2$ over the same set of objects, new kernels may
be built by using  operations such as convex linear
combinations and convolutions. The convolution of $k_1$ and $k_2$ is a
new kernel $k$ with the form

\[ k_1\star  k_2 (u, v) =  \sum_{\{u1 ,u2 \}=u;\{v1 ,v2 \}=v} k_1 (u_1 , v_1 )k_2 (u_2 , v_2 )
\]
where $u = \{u_1 , u_2 \}$ refers to a partition of $u$ into two
substructures $u_1$ and $u_2$
\cite{haussler99convolution,Scholkopf2001}.  The kind of substructures
depends on the domain of course and could be, for instance, subgraphs
or subsets or substrings in the case of kernels defined over graphs,
sets or strings, respectively.  Different kernels can be obtained by
considering different classes of subgraphs (e.g. directed/undirected,
labeled/unlabeled, paths/trees/cycles, deterministic/random walks) and
various ways of listing and counting them
\cite{KashimaTI03,Bunke2006,Neuhaus2007}.  The consideration of space
and time complexity so as to compute convolution/spectral kernels is
important, owing to the combinatorial explosion linked to
variable-size substructures.

  In the following section we present our {\it Optimal Assignment Kernel}
as a symmetric and positive definite similarity measure for directed
graph structures and it will be used in order to define the correspondence
between the vertices of two directed graphs.
For an introduction to kernel functions related concepts and notation, the
reader is referred to  Scholkopf and Smola's book \cite{Scholkopf2001}.

\section{Case-Based Planning and \textsc{OAKplan}}
\label{sec:CaseBasedPlanning}

Here we provide a detailed description of the case-based approach to
planning and its implementation in the state-of-the-art CBP system
\textsc{OAKplan}.\\ \\
A {\it case-based planning system} solves planning problems by making use of
stored plans that were used to solve analogous problems. CBP is a type of
case-based reasoning, which involves the use of stored experiences
({\it cases}); moreover there is strong evidence that people frequently employ
this kind of analogical reasoning  \cite{Gentner93,Ross89,VosniadouOrtony89}.
When a CBP system solves a new planning problem, the new plan is added
to its case  base for potential reuse in the future. Thus we can say
that the system learns from experience.

In general the following steps are executed when a new planning problem must
be solved by a CBP system:
\begin{enumerate}

\item {\it Plan Retrieval}  to retrieve cases from memory that are analogous to the
  current ({\it target}) problem  (see section \ref{sec:retrieval} for a description of our approach).

\item {\it  Plan Evaluation} to evaluate the new plans by execution, simulated
  execution, or analysis and choose one of them (see section \ref{sec:plan_eval}).

\item {\it  Plan Adaptation} to repair any faults found in the new plan (see section \ref{sec:adaptation}).

\item {\it  Plan Revision}  to test the solution new plan $\pi$ for success and repair it if a failure
  occurs during execution (see section \ref{sec:revision}).

\item  {\it Plan Storage}  to eventually store $\pi$ as a new case in the case  base (see section \ref{storingphase}).
\end{enumerate}

 \begin{figure*}[tbp]
 \begin{center}
 \includegraphics[angle=0,width=0.6\textwidth]{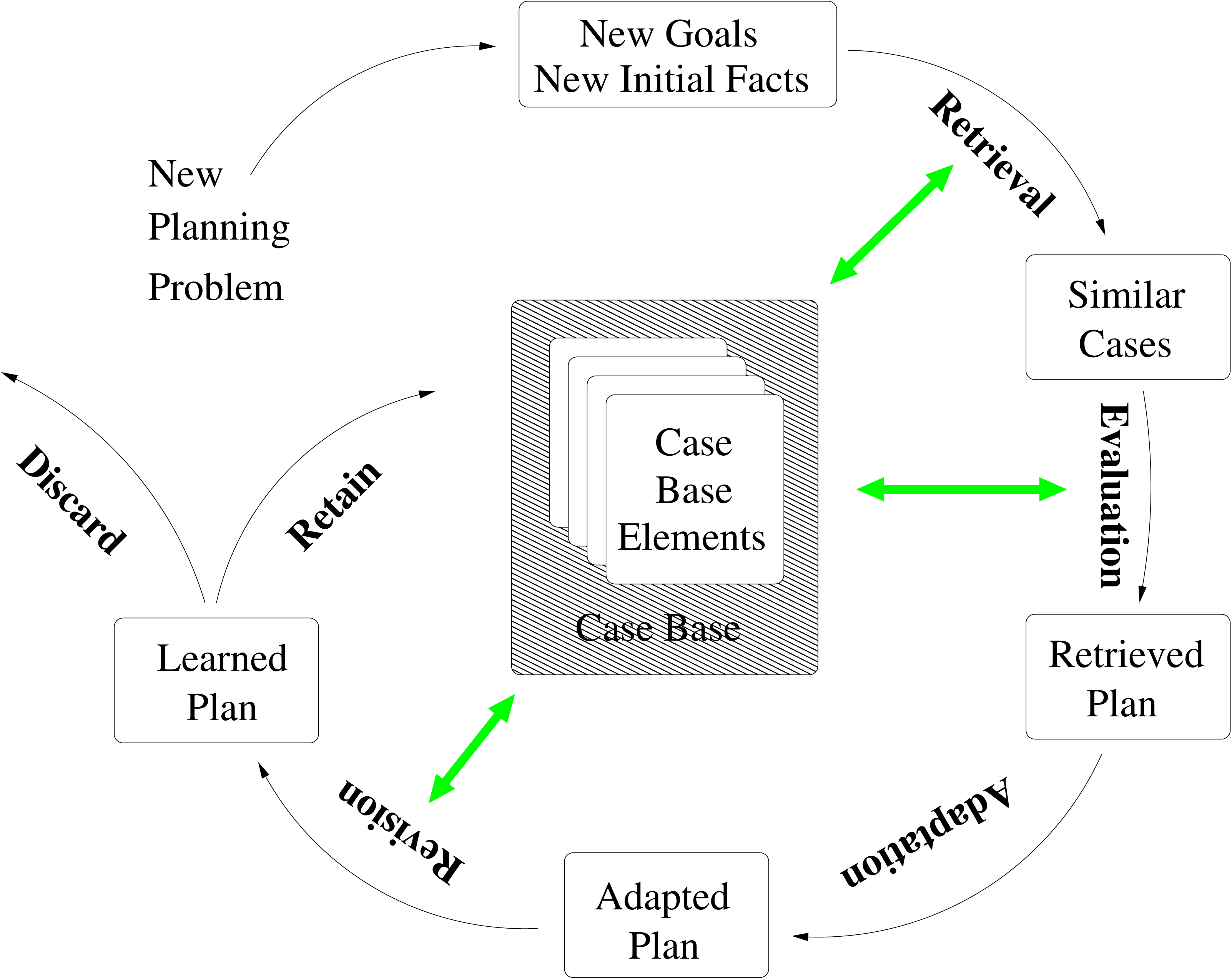}
 \end{center}
 \vspace{-4mm}
 \caption{\ifTR \small \fi \label{fig:cbp-cycle}
 The case-based planning cycle.
 }
 \vspace{-2mm}
 \end{figure*}

 In order to realise the benefits of remembering and reusing past
 plans, a CBP system needs efficient methods for retrieving analogous
 cases and for adapting retrieved plans together with a case  base of
 sufficient size and coverage to yield useful analogues.
 The ability of the system to search in the library for a plan
 suitable to adaptation\footnote{A plan suitable to adaptation has an
   adaptation cost that is lower with respect to the other candidates
   of the case base and with respect to plan generation.} depends both
 on the efficiency/accuracy of the implemented retrieval algorithm and
 on the data structures used to represent the elements of the case
 base.

\ifTR

A planning case of the case   base corresponds to a planning problem
$\Pi$ (defined by an initial state $I$, a goal state $G$ and a set of
operators $O$) a solution $\pi$ of $\Pi$ and additional data
structures derived by $\Pi$ and stored in the case   base so as to avoid
their recomputation.
 The case   base competence should increase
 progressively during the use of the case        base system itself, every
 time solution plans enhancing the competence of the case   base are
 produced.

The possibility of solving a large number of problems depends both on the size
and on the competence of the library with respect to the agent activity.
Furthermore this competence could be increased during the agent activity, in
fact the solution plans of new problems could be added to the library.

\fi

Similarly to the Aamodt \& Plaza's classic model of the problem
solving cycle in CBR \cite{Aamodt94}, Figure \ref{fig:cbp-cycle} shows
the main steps of our case-based planning cycle and the interactions
of the different steps with the case        base.  In the following we
illustrate the main steps of our case-based planning approach,
examining the different implementation choices adopted.

\subsection{Plan Retrieval}\label{sec:retrieval}

Although the plan adaptation phase is the central component of a CBP
system, the retrieval phase critically affects the system performance
too. As a matter of fact the retrieval time is a component of the {\it
  total adaptation time} and the {\it quality} of the retrieved plan
is fundamental for the performance of the successive adaptation phase.
With {\OAKplan} a number of functions for the management of the plan
library and matching functions for the selection of the candidate plan
for adaptation have been implemented.

The retrieval phase has to consider all the elements of the plan
library in order to choose a good one that will allow the system to
solve the new problem easily.  Hence it is necessary to design a similarity
metric and reduce the number of cases that must be evaluated
accurately so as to improve the efficiency of the retrieval phase.
Anyway the efficiency of a plan adaptation system is undoubtedly
linked to the {\it distance} between the problem to solve and the plan
to adapt. In order to find a plan which is {\it useful} for adaptation
we have to reach the following objectives:

 \begin{itemize}

 \item The retrieval phase must identify the candidates for adaptation. The
 retrieval time  should be as small as possible as it will be added to the
 adaptation time and so particular attention has been given to the creation of
 efficient data structures for this phase.

\item The selected cases should actually contain the plans that are
  easier to adapt; since we assume that the world is regular,
  i.e. that similar problems have similar solutions, we look for the
  cases that are the most similar to the problem to solve (with
  respect to all the other candidates of the case base). In this
  sense, it is important to define a metric able to give an accurate
  measure of the similarity between the planning problem to solve and
  the cases of the plan library.
 \end{itemize}

 To the end of applying the reuse technique, it is necessary to
 provide a plan library from which ``sufficiently similar'' reuse
 candidates can be chosen. In this case, ``sufficiently similar''
 means that reuse candidates have a large number of initial and
 goal facts in common with the new instance. However, one may also
 want to consider the reuse candidates that are similar to the new
 instance after the objects of the selected candidates have been
 systematically renamed. As a matter of fact, every plan reuse system
 should contain a matching component that tries to find a mapping
 between the objects of the reuse candidate and the objects of the new
 instance such that the number of common goal facts is maximised and
 the additional planning effort to achieve the initial state of the
 plan library is minimised.  Following Nebel \& Koehler's
 formalisation \cite{Nebel95}, we will have a closer look at this
 matching problem.

 \subsubsection{Object Matching}
\label{sec:ObjectMatching}

As previously said we use {\it a many-sorted logic} in order to reduce the
search space for the matching process; moreover we assume that the operators
are ordinary STRIPS operators using
\ifTR
 variables, i.e. we require that if there
exists an operator $o_k$ mentioning the typed constants $\{c_1:t_1, ...,
c_n:t_n\} \subseteq {\bf O}$, then there also exists an operator $o_l$ over
the arbitrary set of typed constants $\{d_1:t_1, ..., d_n:t_n\} \subseteq {\bf
  O}$ such that $o_l$ becomes identical to $o_k$ if the $d_i$'s are replaced
by $c_i$'s.
\fi
If there are two instances
\[
 \Pi' =\langle {\Prop}( {\bf O}', {\bf P}'),{\cal I', G'}, \nOp \rangle
\ifTR
 \]
 \[
\fi
\Pi = \langle {\Prop}( {\bf O}, {\bf P}),{\cal I, G}, \Op \rangle
 \]
 such that (without loss of generality)
 \[
 {\bf O}' \subseteq {\bf O}
\ifTR
 \]
 \[
\fi
 {\bf P}'={\bf P}
\ifTR
 \]
 \[
\fi
 {\nOp}\subseteq {\Op}
 \]
 then a {\it mapping}, or {\it matching function}, from $\Pi'$ to $\Pi$ is a function
 \[ \mu : {\bf O}' \rightarrow {\bf O}  \]

The mapping is extended to ground atomic formulae and sets of such formulae in
the canonical way, i.e.,
 \[\mu (p(c_1:t_1,...,c_n:t_n)) =
 p(\mu(c_1):t_1,...,\mu(c_n):t_n)
 \]\[
 \mu \large(\{p_1(..),..., p_m(..)\}\large) = \{\mu(p_1(..)),..., \mu(p_m(..))\}
 \]

 If there exists a bijective matching function $\mu$ from $\Pi'$ to $\Pi$ such that $\mu(G')=G$ and $\mu(I')=I$, then it is obvious that a solution plan
 $\pi'$ for
 $\Pi'$ can be directly reused for solving $\Pi$ since $\Pi'$ and $\Pi$ are
 identical  within a renaming of constant symbols, i.e., $\mu (\pi')$ solves
 $\Pi$. Even if $\mu$ does not match all goal
 and initial-state facts, $\mu(\pi')$ can still be used as a starting point
 for the adaptation process that can solve $\Pi$.

 In order to measure the similarity between two objects, it is
 intuitive and usual to compare the features which are common to both
 objects \cite{Lin98}.  The Jaccard similarity coefficient used in
 information retrieval is particularly interesting. Here we examine an
 extended version that considers two pairs of disjoint sets:
\begin{equation}
complete\_simil_\mu(\Pi', \Pi)=\frac{|\mu( {\cal G}')\cap
 {\cal G}| + |\mu({\cal I}') \cap {\cal I}|}{ |\mu( {\cal G}')\cup
 {\cal G}| + |\mu({\cal I}') \cup {\cal I}|}
\end{equation}

In the following we present a variant of the previous function so as
to overcome the problems related to the presence of irrelevant facts
in the initial state description of the current planning problem $\Pi$
and additional goals that are present in $\Pi'$.  In fact while the
irrelevant facts can be filtered out from the initial state
description of the case-based planning problem $\Pi'$ using the
corresponding solution plan $\pi'$, this is not possible for the
initial state description of the current planning problem
$\Pi$. Similarly, we do not want to consider possible ``irrelevant''
additional goals of $G'$; this could happen when $\Pi'$ solves a more
difficult planning problem with respect to $\Pi$.  We define the
following similarity function so as to address these issues:
\begin{equation}
 simil_\mu(\Pi', \Pi)=\frac{|\mu( {\cal G}')\cap
 {\cal G}| + |\mu({\cal I}') \cap {\cal I}|}{ | {\cal G}| + |\mu({\cal
I}')|}.
 \end{equation}

 Using $simil_\mu$ we obtain a value equal to 1 when there exists a
 mapping $\mu$ s.t.  $ \forall f\in I', \; \mu(f)\in I$ (to guarantee
 the applicability of $\pi'$) {\it and } $\forall g\in G, \; \exists
 g'\in G'$ s.t. $ g=\mu(g')$ (to guarantee the achievement of the
 goals of the current planning problem).  \ifTR Note that these
 similarity functions are not metric functions, although we could
 define a distance function in terms of the similarity as
 $Dist_\mu(\Pi', \Pi)=1 - simil_\mu(\Pi', \Pi) $.  and it easy to show
 that this distance function is indeed a metric.

\fi

Finally we define the following optimisation problem, which we call \OBJMATCH:

\noindent
 {\it Instance:} Two planning instances, $\Pi'$ and $\Pi$, and a real number $k\in [0,1]$.
\\
{\it Question:}
\hspace{0mm}\begin{minipage}[t]{12cm} Does a mapping $\mu$ from $\Pi'$
  to $\Pi$ such that $ simil_\mu(\Pi', \Pi)=k$ exist and there is no
  mapping $\mu'$  from $\Pi'$ to $\Pi$ with $simil_{\mu'}(\Pi', \Pi) >
  k $?\\
\end{minipage}

It should be noted that this matching problem has to be solved for
each potentially relevant candidate in the plan library to select the
corresponding best reuse candidate. Of course, one may use structuring
and indexing techniques to avoid considering all plans in the
library. Nevertheless, it seems unavoidable solving this problem a
considerable number of times before an appropriate reuse candidate is
identified. For this reason, the efficiency of the matching component
is crucial for the overall system performance. Unfortunately,
similarly to Nebel \& Koehler's analysis \cite{Nebel95}, it is quite
easy to show that this matching problem is an NP-hard problem.

 \begin{theorem} \label{th:obj_match}
 {\OBJMATCH} is NP-hard.
 \end{theorem}

 The proof of this theorem and of the following ones can be found in
 Appendix \ref{sec:proofs}. This NP-hardness result implies that
 matching may be indeed a bottleneck for plan reuse systems. As a
 matter of fact, it seems to be the case that planning instances with
 complex goal or initial-state descriptions may not benefit from
 plan-reuse techniques because matching and retrieval are too
 expensive.   In fact existing similarity metrics address the
   problem heuristically, considering approximations of it
   \cite{Munoz-Avila96,veloso92learning}. However, this theorem
 is interesting because it captures the limit case for such
 approximations.

\ifTR
\paragraph{Planning Encoding Graph.}
\fi

 We define a particular labeled graph data structure called {\it
   Planning Encoding Graph} which encodes the initial and goal facts
 of a single planning problem $\Pi$ to perform an efficient matching
 between the objects of a planning case and the objects of the current
 planning problem.  The vertices of this graph belong to a set ${\bf
   V}_\Pi$ whose elements are the representation of the objects {\bf
   O} of the current planning problem $\Pi$ and of the predicate
 symbols {\bf P} of $\Pi$:
\[ {\bf V}_\Pi = {\bf O} \cup \bigcup_{p\in{\bf P}} I_p \cup \bigcup_{q\in{\bf
        P}} G_q
\]
i.e. for each predicate we define two additional nodes, one associated
to the corresponding initial fact predicate called $I_p$ and the other
associated to the corresponding goal fact predicate called $G_q$. The
labels of this graph are derived from the predicates of our facts and
the sorts of our many-sorted logic. The representation of an entity
(an object using planning terminology) of the application domain
is traditionally called a {\em concept} in the conceptual graph
community \cite{Chein2008}. Following this notation a {\it Planning Encoding
Graph} is composed of three kinds of nodes: {\em concept} nodes
representing entities (objects) that occur in the application domain,
{\em initial fact relation} nodes representing relationships that hold
between the objects of the initial facts and {\em goal fact relation}
nodes representing relationships that hold between the objects of the
goal facts.

 \begin{figure*}[tbp]
 \begin{center}
 \includegraphics[angle=0,width=0.8\textwidth]{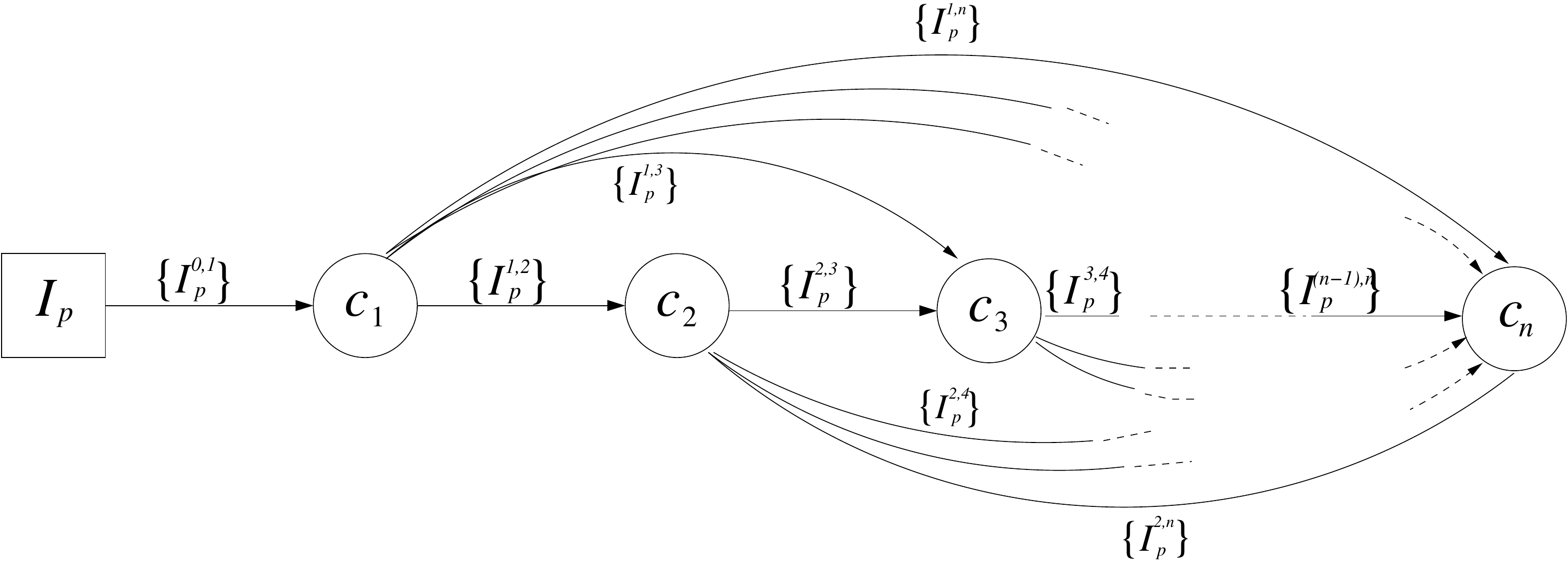}
\[
\lambda(I_p)=\{(I_p,1)\}=\{I_p\},~~ \lambda(c_1)=\{(t_1,1)\}=\{t_1\},
..., \lambda(c_n)=\{(t_n,1)\}=\{t_n\}
\]
 \end{center}
 \vspace{-1mm}
 \caption{\ifTR \small \fi \label{fig:fact-encoding}
 Initial Fact Encoding Graph $\EG^I({\bf p})$ of the propositional
 initial fact ${\bf p}=p(c_1 : t_1, ..., c_n : t_n)$ }


 \begin{center}
 \includegraphics[angle=0,width=0.6\textwidth]{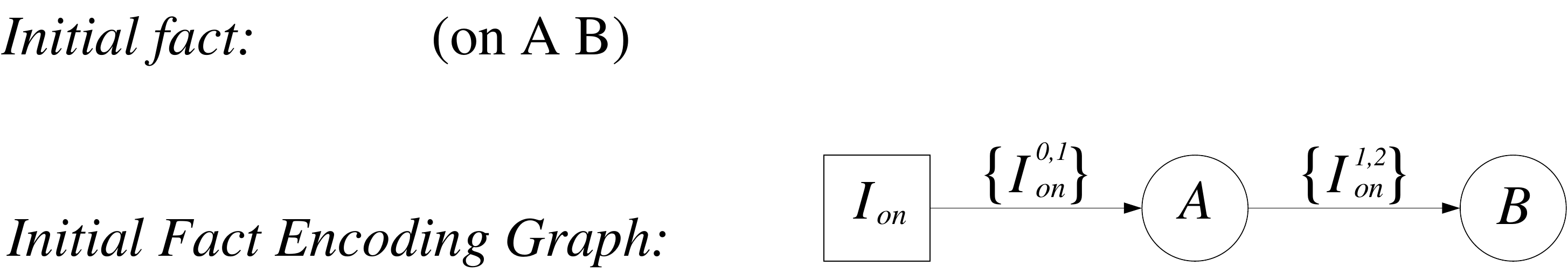}
\[
\lambda(I_{on})=\{I_{on}\},~~ \lambda(A)=\{Obj\},~~
 \lambda(B)=\{Obj\}
\]
 \end{center}
 \vspace{-1mm}
 \caption{\ifTR \small \fi \label{fig:fact-on_A_B}
 Initial Fact Encoding Graph $\EG^I( on~ A~ B )$ of the propositional
 initial fact $(on~ A~ B)$. }

 \end{figure*}

 The {\it Planning Encoding Graph} of a planning problem $\Pi(I,G)$ is
 built using the corresponding initial and goal facts. In particular
 for each propositional initial fact ${\bf p}=p(c_1 : t_1, ..., c_n :
 t_n)\in I$ we define a data structure called {\it Initial Fact
   Encoding Graph} which corresponds to a graph that represents ${\bf
   p}$. More precisely:
 \begin{definition} \label{def:IFEG} Given a propositional typed
   initial fact ${\bf p}=p(c_1 : t_1, ..., c_n : t_n)\in I$ of $\Pi$,
   the {\bf Initial Fact Encoding Graph} $\EG^I({\bf p})=(V_{\bf p},
   E_{\bf p}, \lambda_{\bf p})$ of fact ${\bf p}$ is a directed
   labeled graph where
 \begin{itemize}
 \item $V_{\bf p}=\{I_p, c_1, ... , c_n\}\subseteq {\bf V}_\Pi$;
 \item $E_{\bf p}= \{[I_p,c_1],[c_1, c_2],[c_1, c_3], ..., [c_1, c_n], [c_2, c_3],[c_2, c_4],  ...,[c_{n-1},c_n] \}=$
\[=[I_p,c_1] \cup \bigcup_{i=1,\dots,n;~j=i+1,\dots,n }[c_i, c_j] \]
 \item $\lambda_{\bf p}(I_p)=\{I_p\}, \lambda_{\bf p}(c_i)=\{t_i\} {\tt ~with~}
 {i=1,...,n}$;
 \item        $\lambda_{\bf p}([I_p,c_1])=\{I^{0,1}_p\} $; ~~~~ $\forall [c_i, c_j]\in E_{\bf p}, ~ \lambda_{\bf p}([c_i, c_j])=\{I^{i,j}_p\} $;
 \end{itemize}
 \end{definition}

 i.e. the first node of the graph $\EG^I({\bf p})$, see Figure
 \ref{fig:fact-encoding}, is the initial fact relation node $I_p$
 labeled with the multiset $\lambda_{\bf
   p}(I_p)=\{(I_p,1)\}=\{I_p\}$,\footnote{In the following we indicate
   the multiset $\{(x,1)\}$ as $\{x\}$ for sake of simplicity.} it is
 connected to a direct edge to the second node of the graph, the
 concept node $c_1$, which is labeled by sort $t_1$
 $  \Large\left( \normalsize\textrm{i.e. }\lambda_{\bf p}(c_1)=\{(t_1,1)\}=\{t_1\} \Large\right)$;
 the node $c_1$ is connected with the third node of the graph $c_2$
 which is labeled by sort $t_2$ $\Large\left(\normalsize\textrm{i.e. } \lambda_{\bf
   p}(c_2)=\{(t_2,1)\}=\{t_2\}\Large\right)$ and with all the remaining concept
 nodes, the third node of the graph $c_2$ is connected with $c_3$,
 $c_4$, ...,$c_n$ and so on.  The first edge of the graph $[I_p,c_1]$
 is labeled by the multiset $\{I^{0,1}_p,1\}=\{I^{0,1}_p\}$, similarly
 a generic edge $[c_i, c_j]\in E_{\bf p}$ is labeled by the multiset
 $\{I^{i,j}_p\} $.

 For example, in Figure \ref{fig:fact-on_A_B} we can see the Initial
 Fact Encoding Graph of the fact ``${\bf p}=(on~ A~ B)$'' of the
 BlocksWorld domain. The first node is named as ``$I_{on}$'' and its
 label is the multiset $\lambda_{\bf
   p}(I_{on})=\{(I_{on},1)\}=\{I_{on}\}$, the second node represents
 the object ``A'' with label $ \lambda_{\bf p}(A)=\{(Obj,1)\}=\{Obj\}$
 and finally the third node represents the object ``B'' and its label
 is $ \lambda_{\bf p}(B)=\{Obj\}$; the label of the $[I_{on},A]$ arc
 is the multiset $\{(I^{0,1}_{on},1)\}=\{I^{0,1}_{on}\}$ and the label
 of the $[A,B]$ arc is the multiset
 $\{(I^{1,2}_{on},1)\}=\{I^{1,2}_{on}\}.$

 Similarly to Definition \ref{def:IFEG} we define the {\bf Goal Fact
   Encoding Graph} $\EG^G({\bf q})$ of the fact ${\bf q}=q(c'_1 :
 t'_1, ..., c'_m : t'_m)\in G$ using $\{G_q\}$ for the labeling
 procedure.

 \begin{figure*}[tbp]
 \begin{center}
 \includegraphics[angle=0,width=0.6\textwidth]{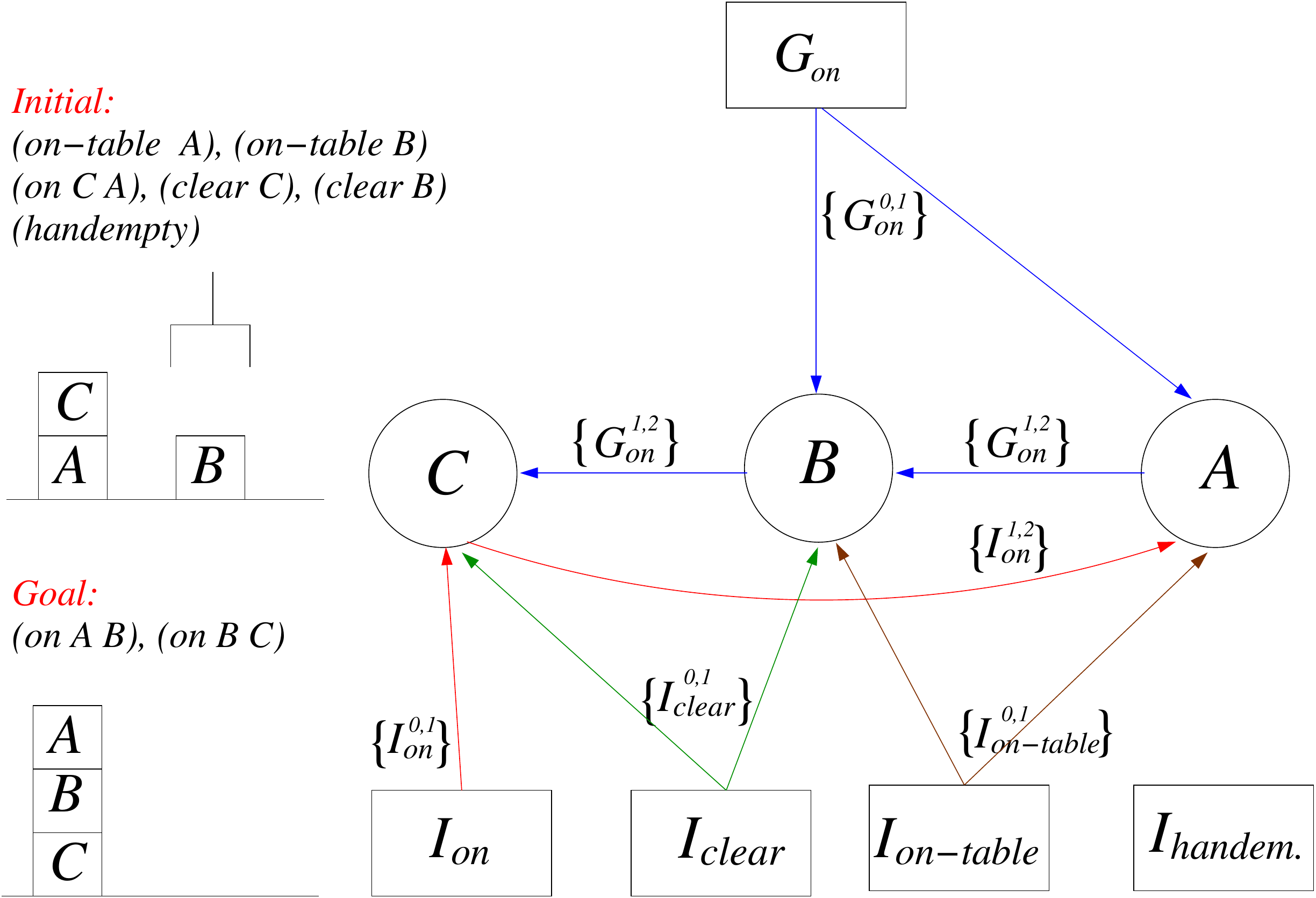}

\vspace{5mm}

$\lambda(A)=\{(Obj, 3)\}$, $\lambda(B)=\{(Obj, 4)\}$ and $\lambda(C)=\{(Obj,
3)\}$
 \end{center}
 \vspace{-1mm}
 \caption{\ifTR \small \fi \label{fig:matching-graph} Planning Encoding Graph for the Sussman Anomaly
 planning problem in the BlocksWorld domain. }

 \end{figure*}

 Given a planning problem $\Pi$ with initial and goal states $I$ and
 $G$, the {\bf Planning Encoding Graph} of $\Pi$, that we indicate as
 ${\PEG}_{\Pi}$, is a directed labeled graph derived by the encoding
 graphs of the initial and goal facts:
 
\begin{equation}
{\PEG}_{\Pi(I,G)}= \bigcup_{{\bf p}\in I} \EG^I({\bf p}) \cup \bigcup_{{\bf q}\in G} \EG^G({\bf q})  
\end{equation}

i.e. the { Planning Encoding Graph} of $\Pi(I,G)$ is a graph obtained
by merging the Initial and Goal Fact Encoding Graphs.
For simplicity in the following we visualise it as a three-level
graph. The first level is derived from the predicate symbols of the
initial facts, the second level encodes the objects of the initial and
goal states and the third level shows the goal fact nodes
 derived from the predicate symbols of the goal
facts.\footnote{ Following the conceptual graph notation, the first
         and third level nodes correspond to initial and goal fact relation
         nodes, while the nodes of the second level correspond to concept
         nodes representing the objects of the initial and goal states.}

Figure \ref{fig:matching-graph} illustrates the Planning Encoding
Graph for the Sussman anomaly planning problem in the BlocksWorld domain.
The nodes of the first and third levels are the initial and goal fact
relation nodes: the vertices $I_{on}$, $I_{clear}$ and $I_{on-table}$
are derived by the predicates of the initial facts, while $G_{on}$ by
the predicates of the goal facts. The nodes of the second level are concept nodes which
represent the objects of the current planning problem $A$, $B$ and $C$,
where the label ``$Obj$'' corresponds to their type. The initial
fact ``{\it (on~C~A)}'' determines two arcs, one connecting $I_{on}$
to the vertex $C$ and the second connecting $C$ to $A$; the labels of
these arcs are derived from the predicate symbol ``$on$'' determining
the multisets $\{I^{0,1}_{on}\}$ and $\{I^{1,2}_{on}\}$ respectively.
In the same way the other arcs are defined. Moreover since there is no
overlapping among the edges of the Initial and Goal Fact Encoding
Graphs, the multiplicity of the edge label multisets is equal to $1$;
on the contrary the label multisets of the vertices associated to the
objects are:

 $\lambda(A)=\{(Obj, 3)\}$, $\lambda(B)=\{(Obj, 4)\}$
and $\lambda(C)=\{(Obj, 3)\}$. \vspace{1mm}

\ifTR
Moreover it could be useful to point out that if an object $c$ appears
more than once in an initial (goal) fact $p(c_1....c_n)$ of a planning
problem $\Pi$, then the corresponding
Initial (Goal) Fact Encoding Graph is built as usual (instantiating $n$
nodes, one each $c_i$), while during the construction of the Planning
Encoding Graph obtained by merging the Initial and Goal Encoding
Graphs of $\Pi$, the nodes that correspond to
the same object are merged into a single vertex node.
\fi

This graph representation can give us a detailed description of the
``topology'' of a planning problem without requiring any a priori assumptions
on the relevance of certain problem descriptors for the whole graph.
Furthermore it allows us to use Graph Theory based techniques in order to
define effective matching functions. In fact a matching function from $\Pi'$
to $\Pi$ can be derived by solving the {\it Maximum Common Subgraph} problem
on the corresponding Planning Encoding Graphs.  A number of exact and approximate
algorithms have been proposed in the literature so as to solve
this graph problem  efficiently.
With respect to {\em normal conceptual graphs} \cite{Chein2008} used
for Graph-based Knowledge Representation, we use a richer label
representation based on multisets. A single relation node is
used to represent each predicate of the initial and goal facts which
reduces the total number of nodes in the graphs considerably. This is
extremely important from a computational point of view since, as we
will see in the following sections, the matching process must be
repeated several times and it directly influences the total retrieval
time.

In the following we examine a procedure based on graph {\it degree
  sequences} that is useful to derive an upper bound on the size of
the {\it MCES} of two graphs in an efficient way. Then we present an
algorithm based on {\it Kernel Functions} that allows to compute an
approximate matching of two graphs in polynomial time.

 \begin{figure}[tb]
\small
\hrule
\vspace{1mm}
 \begin{center}
 \begin{tabbing}
{\it Algorithm} {\sc RelaxedPlan}\\
  \emph{Input}: a set of goal facts ($G$), the set of facts that are true in
  the current state
({\em INIT}),
\\
  a  possibly empty relaxed plan ($A$) \\
  \emph{Output}: a relaxed plan $ACTS$ estimating a minimal set  
of actions required to achieve $G$ \\
1.\hspace{4mm}\= $ G = G -INIT$; $ACTS = A$
\\
2. \> $F = \bigcup_{a \in ACTS} Add(a)$\\
3.\>{\tt while} $G - F \neq \emptyset$
\\
4.\>\hspace{5mm}\=  $g =$ ``a fact in $G - F$''
\\
5.\>\>$bestact = Bestaction(g)$
\\
6. \>\> $Rplan =$ {\sf RelaxedPlan}$\Large\left( Pre(bestact),INIT,ACTS \Large\right)$
\\
7.\>\>  $ACTS = Aset(Rplan) \cup \{ bestact \}$
\\
8.\>\>  $F = \bigcup_{a \in ACTS} Add(a)$
\\
9.\>{\tt return} $ACTS$\\
 \end{tabbing}
\vspace{-2mm}
\hrule
\caption{\ifTR \small \fi \label{fig:relaxed_plan} \small Algorithm for computing a relaxed
  plan estimating a minimal set of actions required to achieve a set of facts
  $G$ from the state {\em INIT}.  $Bestaction(g)$ is the action that is
  heuristically chosen to support $g$  as described in
  \cite{GerSaeSer:JAIR03}.  }
 \end{center}
 \end{figure}

 \begin{figure}[tb]
\small
\hrule
\vspace{1mm}
 \begin{center}
 \begin{tabbing}
{\it Algorithm} {\sc EvaluatePlan}\\
  \emph{Input}: a planning problem $\Pi=(I,G)$, an input plan $\pi$ and an adaptation
 cost limit $Climit$ \\
  \emph{Output}: a relaxed plan to adapt $\pi$ in order to resolve $\Pi$\\
1.\hspace{4mm}\= {\it CState}= $I$;~~ $Rplan = \emptyset$ \\
2. \>{\tt forall} $a \in \pi_i $ {\tt do}\\
3.\>\hspace{5mm}\= {\tt if} \= $\exists f \in Pre(a)$ s.t $f\not \in {\it CState} $ {\tt ~then~}  \\
4. \>\>\> $Rplan =  ${\sc RelaxedPlan}$\Large\left( Pre(a),CState,Rplan \Large\right)$ \\
5.\>\> {\tt if} \> $|Rplan|>Climit$ {\tt ~then~}  \\
6. \>\>\> {\tt return} $Rplan$\\
7.\>\> ${\it CState}=({\it CState}/ Del(a)) \cup Add(a)$\\
8.\> {\tt if} $\exists g \in G$ s.t $g\not \in {\it CState} $ {\tt ~then~}  \\
\>\> $Rplan =  ${\sc RelaxedPlan}$\Large\left(G,CState,Rplan\Large\right)$ \\
9.\>{\tt return} $Rplan$\\
 \end{tabbing}
\vspace{-2mm}
\hrule
 \caption{\ifTR \small \fi\label{fig:evaluate_plan}
 \small Algorithm to evaluate the ability of $\pi$ to solve the planning
problem $\Pi$
 }
 \end{center}
 \end{figure}

\subsection{Plan Evaluation Phase}\label{sec:plan_eval}

The purpose of plan evaluation is that of defining the capacity of a
plan $\pi$ to resolve a particular planning problem. It is performed
by simulating the execution of $\pi$ and identifying the unsupported
preconditions of its actions; in the same way the presence of
unsupported goals is identified. The plan evaluation function could be
easily defined as the number of inconsistencies in the current
planning problem. Unfortunately this kind of evaluation considers a uniform
cost in order to resolve the different inconsistencies and this
assumption is generally too restrictive. Then our system considers
a more accurate inconsistency evaluation criterion so as to improve
the plan evaluation metric. The inconsistencies related to unsupported
facts are evaluated by computing a relaxed plan starting from the
corresponding state and using the {\sc RelaxedPlan} algorithm in {\sc
  lpg} \cite{GerSaeSer:JAIR03}. The number of actions in the relaxed
plan determines the difficulty to make the selected
inconsistencies supported; the number of actions in the final relaxed
plan determines the accuracy of the input plan $\pi$ to solve the
corresponding planning problem.

 Figure \ref{fig:relaxed_plan} describes the main steps of the {\sc
   RelaxedPlan} function.\footnote{ {\sc RelaxedPlan} is described in
   detail in~\cite{GerSaeSer:JAIR03}. It also computes an estimation
   of the earliest time when all facts in $G$ can be achieved, which
   is not described in this paper for sake of simplicity.  }  It
 constructs a relaxed plan through a backward process where
 $Bestaction(g)$ is the action $a'$ chosen to achieve a (sub)goal $g$,
 and such that: (i) $g$ is an effect of $a'$; (ii) all preconditions
 of $a'$ are reachable from the current state {\em INIT}; (iii) the reachability of the
 preconditions of $a'$ requires a minimum number of actions, evaluated
 as the maximum of the heuristically estimated minimum number of
 actions required to support each precondition $p$ of $a'$ from {\em
   INIT}; (iv) $a'$ subverts a minimum number of supported
 precondition nodes in $\Act$ (i.e., the size of the set {\em
   Threats}$(a')$ is minimal).

 Figure \ref{fig:evaluate_plan} describes the main steps of the {\sc
   EvaluatePlan} function. For all actions of $\pi$ (if any), it
 checks if at least one precondition is not supported. In this case it
 uses the {\sc RelaxedPlan} algorithm (step 4) so as to identify the
 additional actions required to satisfy the unsupported
 preconditions. If $Rplan$ contains a number of actions greater than
 $Climit$ we can stop the evaluation, otherwise we update the current
 state {\it CState} (step 7). Finally we examine the goal facts $G$
 (step 8) to identify the additional actions required to satisfy them,
 if necessary.

In order to improve the efficiency of the system and reuse
as many possible parts of previously executed plans we have adopted
{\em plan merging techniques} \cite{Yang92}, which are based on the well-known {\em divide and conquer} strategy.

 In order to apply this strategy, our system must accomplish two
 further subtasks: {\em problem decomposition} and {\em plan
 merging}. The problem decomposition is performed identifying the set
 of actions and the initial facts needed for a single goal and storing
 them in the case base as a new problem instance (if not already
 present); moreover these new instances remain related to the original
 solution plan in order to maintain a statistic of their effective
 usage.  The stored (sub)cases are then used in the merging phase in
 order to identify a single global plan that satisfies all goals. We
 progressively identify the unsatisfied goals and the corresponding
 (sub)cases that allow to satisfy them, giving the preference to the
 (sub)plans that allow us to improve the plan metric and that have been
 successful in a greater number of times in analogous situations.

 \subsubsection{Application of plan merging techniques}

We have used case-based plan merging techniques to store plans.
  Moreover, in order to reuse as many as possible parts
of previously executed plans, we decompose the solution plans into
subparts that allow us to satisfy every single goal or a set of
interrelated goals and we store these subparts
in the case base, if they are not already present.

When a new e-learning planning problem must be solved, we search in
the case base if a plan that already solves all goals exists. If such
a plan does not exist we apply plan merging techniques that
progressively identify (sub)plans in the case base that can satisfy
the goals.  This phase consists in reusing parts of the
retrieved plans to complete a new one.  Figure \ref{fig:merge_plans}
describes the process for merging plans of the library in order to
find a plan $\pi$ that solves the current planning problem $\Pi$ or
that represents a quasi-solution \cite{Gerevini99} for it.  At step 1 we
search in the library the plan that satisfies all the goals with the
lowest heuristic adaptation cost, where the function ${\sc EvPlan}(I,
\pi, G)$ determines the adaptation effort by estimating the number of
actions that are necessary to transform $\pi$ into a solution of the
problem $\Pi(I,G)$.\footnote{See {\sc EvaluatePlan} for a more detailed description.} This step corresponds to the extraction
of the best plan of the library (if it exists) as proposed by the
standard {\OAKplan} system.  At steps 3.x, we
progressively analyse the unsatisfied goals and the unsatisfied
preconditions of the current plan $\pi$, trying to identify in the
library a subplan $\pi_f$ that can be merged with $\pi$ in order to
satisfy $f$ (and other unsatisfied facts if possible) and reduce, at the
same time, the global heuristic adaptation cost, where ${\sc merge}$
identifies the best part of $\pi$ where the actions of $\pi_f$ can be
inserted in producing a new global plan.\footnote{In our tests we have
  considered the earliest and the latest part of $\pi$ where $f$ can
  be satisfied.} If such a plan exists, we merge it with $\pi$ at step
3.3 and we restart from step 3 reconsidering all the unsatisfied
facts.  The repeat loop halts when all the goals and
preconditions are satisfied, i.e. when we have found a solution plan, or
when there is not a suitable plan that can be extracted from the
library that satisfies the remaining unsupported facts.  In this
case, the plan $\pi$ does not represent a solution plan. However, it
can be used as a starting point for a local search process to
find a solution plan for the current planning problem.

 \begin{figure}[tb]
\small
\hrule
\vspace{1mm}
 \begin{center}
 \begin{tabbing}
{\it Algorithm} {\sc MergeSubplans}\\
\emph{Input}: A planning problem $\Pi(I,G)$, a plan library ${\cal L}=(\Pi_i,\pi)$;\\
 \emph{Output}: A (quasi) solution plan $\pi$ for $\Pi$;\\
1.\hspace{2mm}\= $\pi = {argmin}_{\pi_i\in {\cal L}\cup \emptyset}{\sc EvPlan}(I, \pi_i,G)$;\\
2.\>  {\tt repeat } \\
3.\>  \hspace{2mm}\={\tt forall} unsatisfied facts $f\in \{G \cup prec(\pi)\}$ {\tt do} \\
3.1\> \>\hspace{2mm}\= Let  $\pi_f\in {\cal L}$ be the best plan that satisfies $f$ s.t. \\
\>\>\> ${\sc EvPlan}(I, {\sc merge}(\pi, \pi_f), G) < {\sc EvPlan}(I, \pi, G)$;\\
3.2 \>\>\> {\tt if} $\pi_f\not=\emptyset$ {\tt then}\\
3.3 \>\>\>\hspace{4mm} $\pi={\sc merge}(\pi, \pi_f)$; break; \\
4. \> {\tt until} $\pi_f\not=\emptyset$;\\
5. \>  {\tt return} $\pi$;
 \end{tabbing}

\hrule

\vspace{-2mm}
\caption{\label{fig:merge_plans} \small Algorithm for merging the elements in the library in order to solve a planning problem  $\Pi$.}
 \end{center}
\vspace{-7mm}
 \end{figure}
%
%

 \begin{figure}[tb]
\small
\hrule
\vspace{0mm}
\begin{center}
\begin{threeparttable}[b]
\begin{tabbing}
{\it Algorithm} {\sc RetrievePlan}
\\
\emph{Input}: a planning problem $\Pi$, a case base
$C=\langle \Pi_{i}, \pi_{i} \rangle$
\\
\emph{Output}: candidate plan for the
adaptation phase
\\
1.1.\hspace{4mm}\=$\pi_R=\mbox{{\sc
        Evaluate\_plan}}(\Pi,\textsc{empty\_plan}, \infty)$
\\
1.2. \>Define the set of initial relevant facts  of $\Pi$ using $\pi_R$:
$I_{\pi_R}=I \cap \bigcup_{a\in \pi_R}  pre(a)$
\\
1.3.\> Compute the Planning Encoding Graphs ${\PEG}_{\Pi}$ and  
${\PEG}_{\Pi_R}$ of $\Pi(I,G)$ and $\Pi_R(I_{\pi_R},G)$
\\
\> \hspace{5mm} respectively, and the degree sequences $L^j_{\Pi_R}$
\\
1.4.\> {\tt forall} $\Pi_i \in C$ {\tt do}
\\
1.5.\>\hspace{5mm}\=$ simil_i= simil^{ds}({\PEG}_{\Pi_i},{\PEG}_{\Pi_R})$
\\
1.6.\>\>{\tt push}$((\Pi_i,simil_i \large)) ,queue)$
\\
1.7.\>\>$best\_ds\_simil= max \large(best\_ds\_simil, simil_i \large)$
\\
2.1.\> {\tt forall}  $(\Pi_i,simil_i) \in queue${~~s.t.~~}$
best\_ds\_simil-simil_i \leq limit$  {\tt do}\tnote{*}
\\
2.2.\>\>Load the Planning Encoding Graph ${\PEG}_{\Pi_i}$ and compute the matching function $\mu_{base}$ \\
\>\> using $\K_{base}({\PEG}_{\Pi_i},{\PEG}_{\Pi})$
\\
2.3.\>\>{\tt push}$((\Pi_i,\mu_{base}),queue_1)$
\\
2.4.\>\>$best\_\mu_{base}\_simil= max \large(best\_\mu_{base}\_simil,simil_{\mu_{base}}(\Pi_i,\Pi) \large)$
\\
3.1.\>{\tt forall}  $(\Pi_i,\mu_{base})\in queue_1${~~s.t.~~}$
best\_\mu_{base}\_simil- simil_{\mu_{base}}(\Pi_i,\Pi)  \leq limit$  {\tt do}
\\
3.2.\>\>Compute the matching function $\mu_{\cal N}$ using  $\K_{\cal N}({\PEG}_{\Pi_i},{\PEG}_{\Pi})$
\\
3.3.\>\> {\tt if}~~~\= $simil_{\mu_{\cal N}}(\Pi_i,\Pi) \geq
        simil_{\mu_{base}}(\Pi_i,\Pi)  $ {\tt ~then~} $\mu_i=\mu_{\cal N} $
\\
\>\>{\tt else}  $\mu_i=\mu_{base}$
\\
3.4.\>\> {\tt push}$((\Pi_i,\mu_i),queue_2)$
\\
3.5.\>\> $ best\_simil= max \large(best\_simil, simil_{\mu_i}(\Pi_i,\Pi) \large) $\\
4.1.\>${ best\_cost} = \alpha_G \cdot |\pi_R|$; ~~~ {\tt best\_plan} =\textsc{empty\_plan}
\\
4.2.\>{\tt forall}  $(\Pi_i,\mu_i)\in queue_2$ {~~s.t.~~} $best\_simil-
        simil_{\mu_i}(\Pi_i,\Pi) \leq limit  $ {\tt do}
\\
4.3.\>\>Retrieve $\pi_i$ from  $C$
\\
4.4.\>\>${ cost}_i = | \mbox{{\sc EvaluatePlan}}\large(\Pi,\mu_i(\pi_i), \; {
        best\_cost}\cdot simil_{\mu_i}(\Pi_i,\Pi)\large)| $
\\
4.5.\>\> {\tt if} ${ best\_cost}\cdot simil_{\mu_i}(\Pi_i,\Pi) > { cost}_i${\tt ~then~} \\
4.6.\>\>\> ${ best\_cost} = { cost_i}/ simil_{\mu_i}(\Pi_i,\Pi)$
\\
4.7.\>\>\> {\tt best\_plan} = $\mu_i(\pi_i)$
\\
5.1.\>{\tt return} {\tt best\_plan}

 \end{tabbing}
 \begin{tablenotes}
 \item [*] We limited this
  evaluation to the best $700$ cases of  $queue$.\\
 \end{tablenotes}
\end{threeparttable}
\vspace{-1mm}
\hrule
 \caption{\ifTR \small \fi\label{fig:retrive_best}
 \small Algorithm to find a suitable plan for the adaptation phase from a set of
 candidate cases or the empty plan (in case the
 ``generative'' approach is considered more suitable).
 }\vspace{-9mm}
 \end{center}
 \end{figure}

 Figure \ref{fig:retrive_best} describes the main steps of the
 retrieval phase. We initially compute a relaxed plan $\pi_R$ for
 $\Pi$ (step $1.1$) using the {\sc EvaluatePlan} function on the empty
 plan which is needed so as to define the {\it generation} cost of the
 current planning problem $\Pi$ (step 4.1)\footnote{The $\alpha_G$ coefficient
   gives more or less importance to plan  adaptation vs plan generation; if
   $\alpha_G>1$ then it is more likely to perform plan adaptation than
   plan generation.} and an {\it estimate} of the initial
 state relevant facts (step $1.2$).  In fact we use the relaxed plan
 $\pi_R$ so as to filter out the irrelevant facts from the initial
 state description.\footnote{In the relaxed planning graph analysis
   the negative effects of the domain operators are not considered and
   a solution plan $\pi_R$ of a relaxed planning problem can be
   computed in polynomial time \cite{HofNeb-JAIR01}. } This could be
 easily done by considering all the preconditions of the actions of
 $\pi_R$: \vspace{-3mm}
\[ I_{\pi_R}= I
\cap \bigcup_{a\in \pi_R}  pre(a).  \vspace{-3mm} \]

Then in step $1.3$ the Planning Encoding Graph of the current planning
problem $\Pi$ and the degree sequences that will be used in the
screening procedure are precomputed. Note that the degree sequences
are computed considering the Planning Encoding Graph ${\PEG}_{\Pi_R}$
of the planning problem $\Pi_R(I_{\pi_R},G)$ which uses $I_{\pi_R}$
instead of $I$ as initial state. This could be extremely useful in
practical applications when automated tools are used to define the
initial state description without distinguishing among relevant and
irrelevant initial facts.

Steps $1.4$ -- $1.7$ examine all the planning cases of the case base
so as to reduce the set of candidate plans to a suitable number.
It is important to point out that in this phase it is not
necessary to retrieve the complete planning encoding graphs of the case base
candidates $G_{\Pi'}$ but only their sorted degree sequences
$L_{\Pi'}^i$ which are precomputed and stored in the case base. On the
contrary the  planning encoding graph and the degree sequences of the input
planning problem are only computed in the initial preprocessing phase
(step $1.3$).

All the cases with a similarity value sufficiently
close\footnote{In our experiments we used $limit=0.1$. } to the best
degree sequences similarity value $(best\_ds\_simil)$ are examined
further on (steps $2.1$--$2.4$) using the ${\K}_{base}$ kernel
function.  Then all the cases selected at steps $2.x$ with a
similarity value sufficiently close to the best $simil_{\mu_{base}}$
similarity value $(best\_\mu_{base}\_simil)$ (step 3.1) are accurately
evaluated using the $\K_{\cal N}$ kernel function, while the
corresponding $\mu_{\cal N}$ function is defined at step $3.2$.
In steps $3.3$--$3.5$ we select the best matching function
found for $\Pi_i$ and the best similarity value found until now.

We use the relaxed plan $\pi_R$ in order to define an estimate of the
{\it generation} cost of the current planning problem $\Pi$ (step
4.1). The {\tt best\_cost} value allows to select a good candidate
plan for adaptation (which could also be the empty plan).
This value is also useful during the computation of the adaptation cost
through {\sc EvaluatePlan}, in fact if such a limit is exceeded then
it is wasteful to use CPU time and memory to carry out the estimate
and the current evaluation could be terminated. The computation of the
adaptation cost of the empty plan allows to choose between an {\em
  adaptive} approach and a {\em generative} approach, if no plan gives
an adaptation cost  smaller than the empty plan.

For all the cases previously selected with a similarity value
sufficiently close to $best\_simil$ (step $4.2$) the adaptation cost
is determined (step $4.4$). If a case of the case base determines an
adaptation cost which is lower than {\tt best\_cost}$\cdot
simil_{\mu_{i}}(\Pi_i,\Pi)$ then it is selected as the current best
case and also the {\tt best\_cost} and the {\tt best\_plan} are updated (steps
$4.5$--$4.7$). Note that we store the encoded plan $ \mu_{i}(\pi_i)$
in {\tt best\_plan} since this is the plan that can be used by the
adaptation phase for solving the current planning problem $\Pi$.
Moreover we use the $simil_{\mu_{i}}(\Pi_i,\Pi)$ value in steps $4.4$
-- $4.6$ as an {\em indicator} of the effective ability of the
selected plan to solve the current planning problem maintaining the
original plan structure and at the same time obtaining low distance
values.

\subsection{Plan Adaptation}\label{sec:adaptation}

As previously exposed, the plan adaptation system is a fundamental
component of a case-based planner. It consists in reusing and
modifying previously generated plans to solve a new problem and
overcome the limitation of planning from scratch. As a matter of fact,
in planning from scratch if a planner receives exactly the same
planning problem it will repeat the very same planning operations.  In
our context the input plan is provided by the plan retrieval phase
previously described; but the applicability of a plan adaption
system is more general.  For example the need for adapting a
precomputed plan can arise in a dynamic environment when the execution
of a planned action fails, when the new information changing the
description of the world prevents the applicability of some planned
actions, or when the goal state is modified by adding new goals or
removing existing ones \cite{fox-icaps06,GerSaeSer:JAIR03}.

\ifTR
Different approaches have been considered in the literature for plan
adaptation; strategies vary from attempting to reuse the structure of
an existing plan by constructing bridges that link together the fragments
of the plan that fail in the face of new initial
conditions~\cite{Ham90,Hanks:92,HanWel95,kambh-1990:theorplan:inbook:176},
to more dynamic plan modification
approaches that use a series of plan modification operators to attempt
to repair a plan~\cite{LikhachevK05,KrogtW-icaps05}.
\fi
From a theoretical point of view, in the worst case, plan adaptation
is not more efficient than a complete regeneration of the plan
~\cite{Nebel95} when a conservative adaptation strategy is adopted.
However  adapting an existing plan can be  in practice more
efficient than generating a new one from scratch, and, in addition, this
worst case scenario does not always hold, as exposed
in \cite{DanaNau2002} for the Derivation Analogy adaptation approach.
Plan adaptation can also be more convenient when the
new plan has to be as ``similar'' as possible to the original one.

Our work uses the \lpg-adapt system given its good performance in many
planning domains but other plan adaptation systems could be used as
well. \lpg-adapt\ is a local-search-based planner that modifies plan
candidates incrementally in a search for a flawless candidate.
\ifTR
We describe the main components of the \lpg-adapt system in the following
section.
\fi
It is important to point out that this paper relates to the
description of a new efficient case-based planner and in particular to the
definition of effective plan matching functions, no significant changes were
made to the plan adaptation component (for a detailed description
of it see~\cite{fox-icaps06}).

\subsection{Plan Revision \& Case Base Update}\label{sec:revision}\label{storingphase}

Any kind of planning system that works in dynamic environments has to take
into account failures that may arise during plan generation and execution.
In this respect case-based planning is not an exception;
this capability is called plan revision and it is divided
in two subtasks: evaluation and repair. The evaluation step verifies the
presence of failures that may occur during plan execution when the plan does
not produce the expected result.  When a failure is discovered, the system may
react by looking for a repair or aborting the plan.  In this first hypothesis
the \lpg-adapt system is invoked on the remaining part of the plan; in the
latter hypothesis the system repeats the CBP cycle so as to search a new
solution.

\begin{figure}[tbp]
\small
\hrule
\begin{center}
\begin{tabbing}
  ~~\={\it Algorithm  }{\sc Insert\_Case}($Case$, $\pi$, $\Pi(I,G)$)
  \\
  \>{\it Input}: \= a case base $Case$, a solution plan $\pi$ for  planning problem $\Pi$ with initial
  state $I$ and goal \\
  \> \> state $G$.  
  \\
  \>{\it Output}: insert the planning case in $Case$ if not present.
  \\
  1.\>~~~ \=Define the set of initial state relevant facts $I_\pi$ of $\Pi$
  using the input plan $\pi$
  \\
  2. \>\> Compute the Planning Encoding Graph ${\PEG}_\pi $ of $\Pi_\pi( I_\pi,G)$
  \\
  3.\>\>{\tt for} each case $(\Pi_i, \pi_i) \in Case$
  \\
  4.\>\>~~~\= Compute the matching function $\mu_i$ using  $\K_{\cal N}({\PEG}_{\Pi_i},{\PEG}_\pi)$
  \\
  5.\>\>\>{\tt if} $complete\_simil_{\mu_i}(\Pi_i,\Pi_\pi)=1 \wedge |\pi_i|\leq |\pi|$
  {\tt then}
  \\
  6.\>\>\>~~~~~\={\tt return} FALSE;
  \\
  7.\>\>{\tt enfor}
  \\
  8.\>\>Insert the planning problem $\Pi_\pi(I_\pi,G)$, its solution plan
  $\pi$, the Planning Encoding  Graph ${\PEG}_\pi$
  \\
  \>\>\> and the data structures for the screening procedure
  in $Case$
  \\
  9.\>\>{\tt return} TRUE;
  \\
\end{tabbing}
\hrule
\caption{\ifTR \small \fi \label{fig:case-basegener}
\small High-level description of {\sc Insert\_Case}.
}
\end{center}
\vspace{-7mm}

\end{figure}

After finding the plan from the library and after repairing it with the
\lpg-adapt techniques the solution plan can be inserted into the library or be
discarded.  
\ifTR
The case base maintenance is clearly important for the performance
of the system and different strategies have been proposed in the literature
{\cite{SmythM99a,TonidandelR02}}. Furthermore our attention
has been oriented towards the improvement of the competence of the case base;
a solved planning problem is not added to the case base only if there is a
case that solves the same planning problem with a solution of a better
quality.\footnote{ In our experiments we have considered only the number of
  actions for distinguishing between two plans that solve the same planning
  problem but other and more accurate metrics could be easily added,
  i.e. consider for example actions with not unary costs.}  Such a check has
been introduced to the end of keeping only the best solution plans for certain
kinds of problems in the library as there can be different plans that can
solve the same problems with different sets of actions.

\fi
Figure \ref{fig:case-basegener} describes the main steps of the function used
to evaluate the insertion of a planning problem $\Pi$ solved in the case base.
First of all we compute the set of initial state relevant facts $I_\pi$ using
the input plan $\pi$; this set corresponds to a subset of the facts of $I$
relevant for the execution of  $\pi$.  It can be easily computed, as described
in section \ref{sec:plan_eval}, using the preconditions of the actions in
$\pi$:
\[I_{\pi}= I \cap \bigcup_{a\in \pi} pre(a).\]

 Note that $I_\pi$ identifies all the facts required for the
  execution of the plan $\pi$ and that this definition is consistent
  with the procedure used in the {\sc RetrievePlan} algorithm for the
  relaxed plan $\pi_R$.\ifTR\footnote{We have used this simple definition
        instead of using the causal links in $\pi$ in order to compute the
        set of relevant facts since it allows to obtain slightly better
        performance than the corresponding version based on causal links.}
\fi
  Then we compute the Planning Encoding Graph
${\PEG}_{\pi}$ of the new planning problem $\Pi_\pi( I_\pi,G)$ having
$I_\pi$ as initial state instead of $I$.  At steps $3$--$6$ the
algorithm examines all the cases of the case base and if it finds a
case that solves $\Pi$ with a plan of a better quality with respect to
$\pi$ then it stops and exits. In order to do so we use the similarity
function $complete\_simil_{\mu_i}$, described in section
\ref{sec:ObjectMatching}, which compares all the initial and goal
facts of two planning problems.  Otherwise if there is no case that
can solve $\Pi_\pi$ with a plan of a better quality with respect to
$\pi$ then we insert the solved problem in the case base. As we can
observe at step $8$, a planning case is made up not only by $\Pi_\pi$
and $\pi$, but also other additional data structures are precomputed
and added to the case base so that their recomputation during the {\it
  Retrieval Phase} can be avoided.
\ifTR

Recently, the system was extended with a set of maintenance policies
guided by the cases' similarity, as is described in \cite{AIIA,ICCBR}.
\fi

 \begin{figure}[tb]
\small
\hrule
\vspace{1mm}
 \begin{center}
 \begin{tabbing}
{\it Algorithm} {\sc UpdateLibrary}\\
\emph{Input}: A solution plan $\pi$ for $\Pi=(I,G)$ and a set of facts $F$;\\
 \emph{Output}: Update the plan library inserting new elements obtained\\
 considering  subplans  of $\pi$;\\
1.\hspace{4mm}\= compute the set of causal links ${\cal C}_\pi$ in $\pi$;\\
2.\> $S=G \cup F \cup \{ G_j \subseteq G \mid \bigcap_{g_j \in G_j} \pi_{g_j} \not= \emptyset\}$;\\
3.\> {\tt forall} $G_i \in S$ {\tt do} \\
3.1 \>\hspace{3mm}\= {\sc Check\&Insert}$(\Pi_{G_i}, \pi_{G_i})$;\\
 \end{tabbing}
\vspace{-2mm}
\hrule
\vspace{-2mm}
\caption{\label{fig:extract_plans} \small Algorithm for updating the plan library inserting subplans of a given input plan $\pi$.}
 \end{center}
\vspace{-7mm}
 \end{figure}


Figure \ref{fig:extract_plans} describes the algorithm for
updating the plan library with parts of an input plan
$\pi$. In short, {\sc UpdateLibrary} identifies the subplans of
$\pi$ that can be inserted in the plan library to increase
the {\it competence} of the library in itself
\cite{SmythM01,TonidandelR02}.
Here $\pi_{g}$ represents the subplan of $\pi$ that satisfies
$g$ starting from $I$. Note that it can be easily identified
considering the set of causal links ${\cal C}_\pi$ of $\pi$ computed
at step 1.  In a similar way, it is possible to compute the set of
facts $I_{g}$ that are necessary to apply the actions of
$\pi_{g}$.

At step 2 we identify the set of facts that will be examined for the
insertion in the library. In particular we consider all the goals $G$,
the elements of $F$ and the subsets of {\em interacting} goals $G_i$.
The $F$ set represents a set of facts, different by the input goals,
that could be useful for the following merging phase such as
unsupported facts of a previous adaptation phase. Moreover, the sets of
interacting goals $G_i$ can be easily computed considering the actions
in the subplans $\pi_{g_j}$ that are in common to the different goals.

The {\sc Check\&Insert}$((I_{G_i}, G_i), \pi)$ function (step $3.1$) searches if there not
exists a case-base element $(\Pi_j,\pi_j)$ whose goals and initial
state perfectly match with the current goals and initial state,
respectively. In this case, we insert the current planning problem
$\Pi_{G_i}=(I_{G_i},G_i)$ and its solution plan $\pi_{G_i}$ in the library. Otherwise, we
have to decide whether to insert $(\Pi_{G_i},\pi_{G_i})$ and remove $(\Pi_j,\pi_j)$,
or simply skip the insertion of $(\Pi_{G_i},\pi_{G_i})$. In our tests we have used
an update policy that maintains the plan with the lowest number of
actions, but other policies could be used as well considering, for
example, the plan qualities, their makespan, or the robustness to
exogenous events.
Moreover, {\sc Check\&Insert} ignores too small and
too big plans $\pi$; in fact, a small plan $\pi$ could determine the
inclusion in the library of a high number of very small plan fragments
that have to be considered in the merging phase, while a big plan
$\pi$ could determine the insertion in the library of very big
subplans that are difficult to merge.\footnote{In our tests we
  have used $5 \le |\pi| \le 200$.}

\section{Experimental Results}
\label{sec:experiments}

In this section, we present an experimental study aimed at testing the
effectiveness of {\OAKplan} in a number of standard benchmark domains.

We have experimented with several courses, but here focus on a real, large-size \emph{Moodle} course of around 90 LOs on Natural Sciences inspired on \texttt{http://www.profesorenlinea.cl}. We have created nine initial configurations (with 10, 20\dots 90 fictitious students, respectively), and defined 10 variants per configuration (plus an additional variant for the 90 problem), thus considering 100 planning problems in total (the 91 variants plus the 9 initial configurations). Each variant artificially simulates the changes that may occur during the route execution in an
incremental way. That is, in the first variant some equipment is no
longer available. The second variant maintains these changes and
includes restrictions on the students' availability; and so on for the
other variants.


In addition to \OAKplan\ and our case base planner \OAKplan-merge, we have used two state of the art planners, \sgplan\ and \lpg\footnote{For a further description of these planners see
  \texttt{http://ipc.icaps-conference.org}.}.
All tests were performed on an Intel(R) Xeon(TM) CPU 2.40GHz with 2GB of RAM, and censored after 10 minutes. In our tests, the solution plans (i.e. the learning routes) inserted in the case base were obtained by using the best quality plans generated by \lpg\ and \sgplan\ on the
initial-configuration planning problems used to create the
corresponding variants.

\begin{figure*}[318tbph]

\vspace{-2.1cm}
\begin{tabular}{cc}

\includegraphics[angle=-90,width=0.50\textwidth]{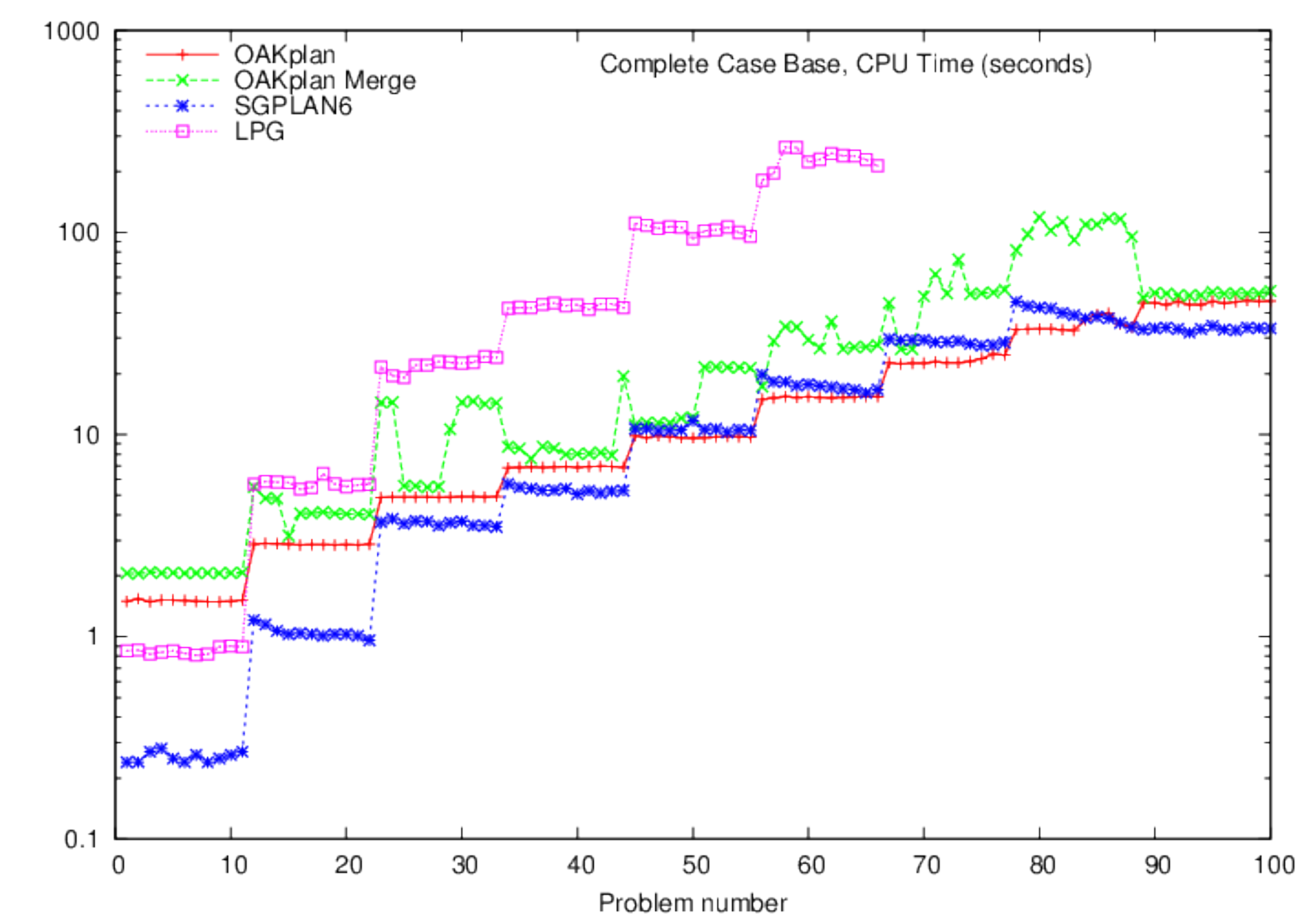}
&
\hspace{-7mm}

\includegraphics[angle=-90,width=0.50\textwidth]{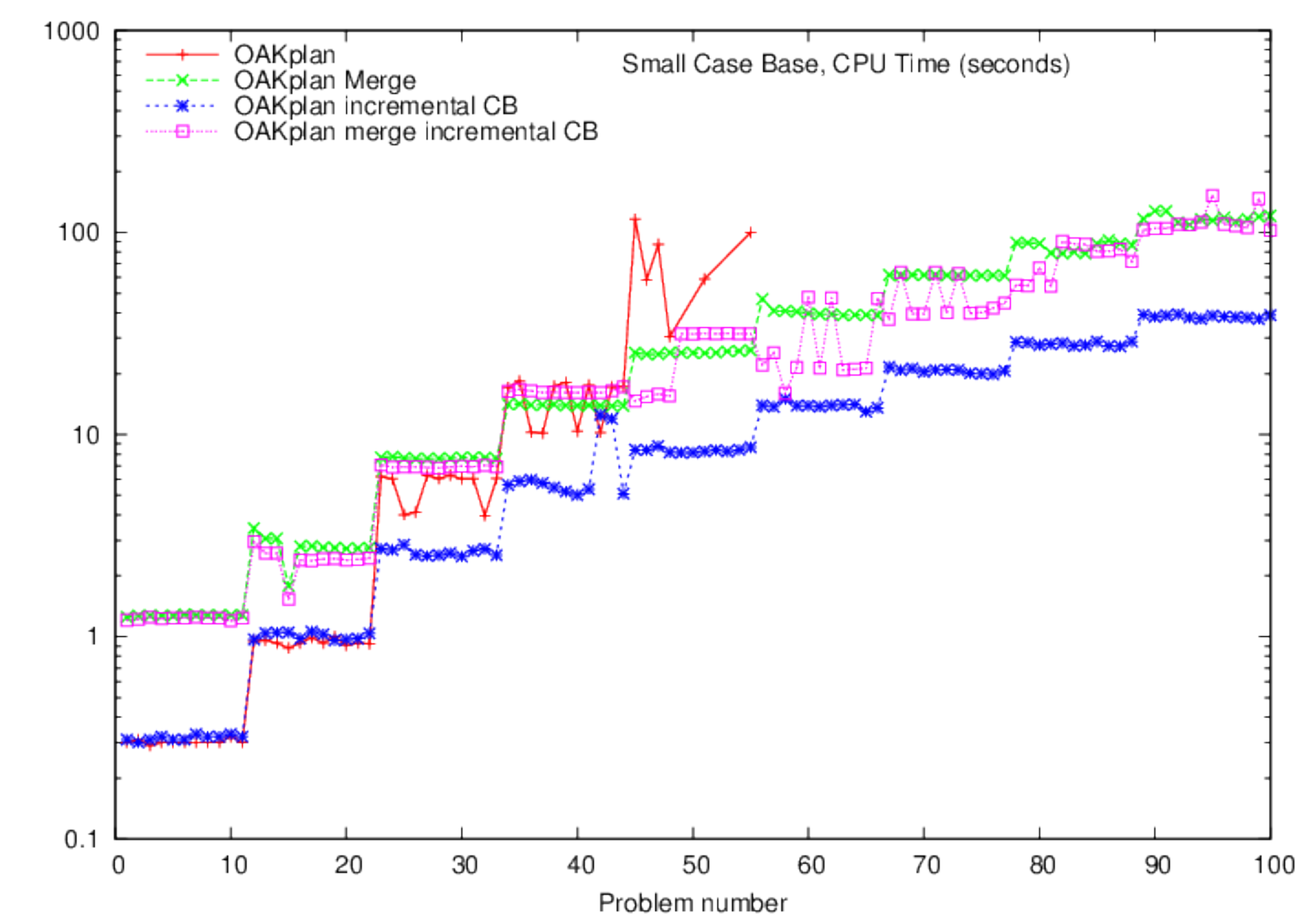}
\vspace{-0.3cm}
\\

\includegraphics[angle=-90,width=0.50\textwidth]{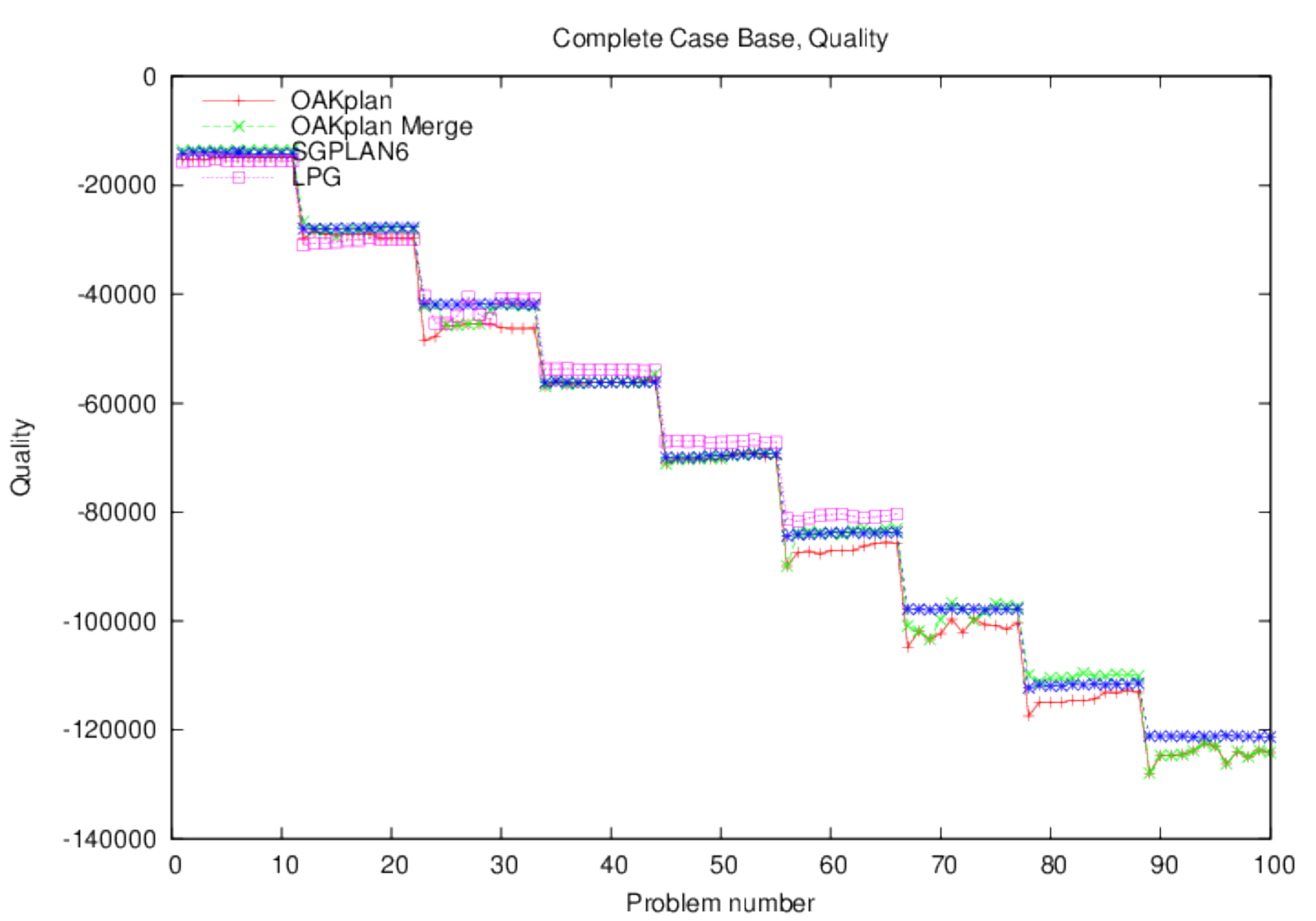}

&
\hspace{-7mm}
\includegraphics[angle=-90,width=0.50\textwidth]{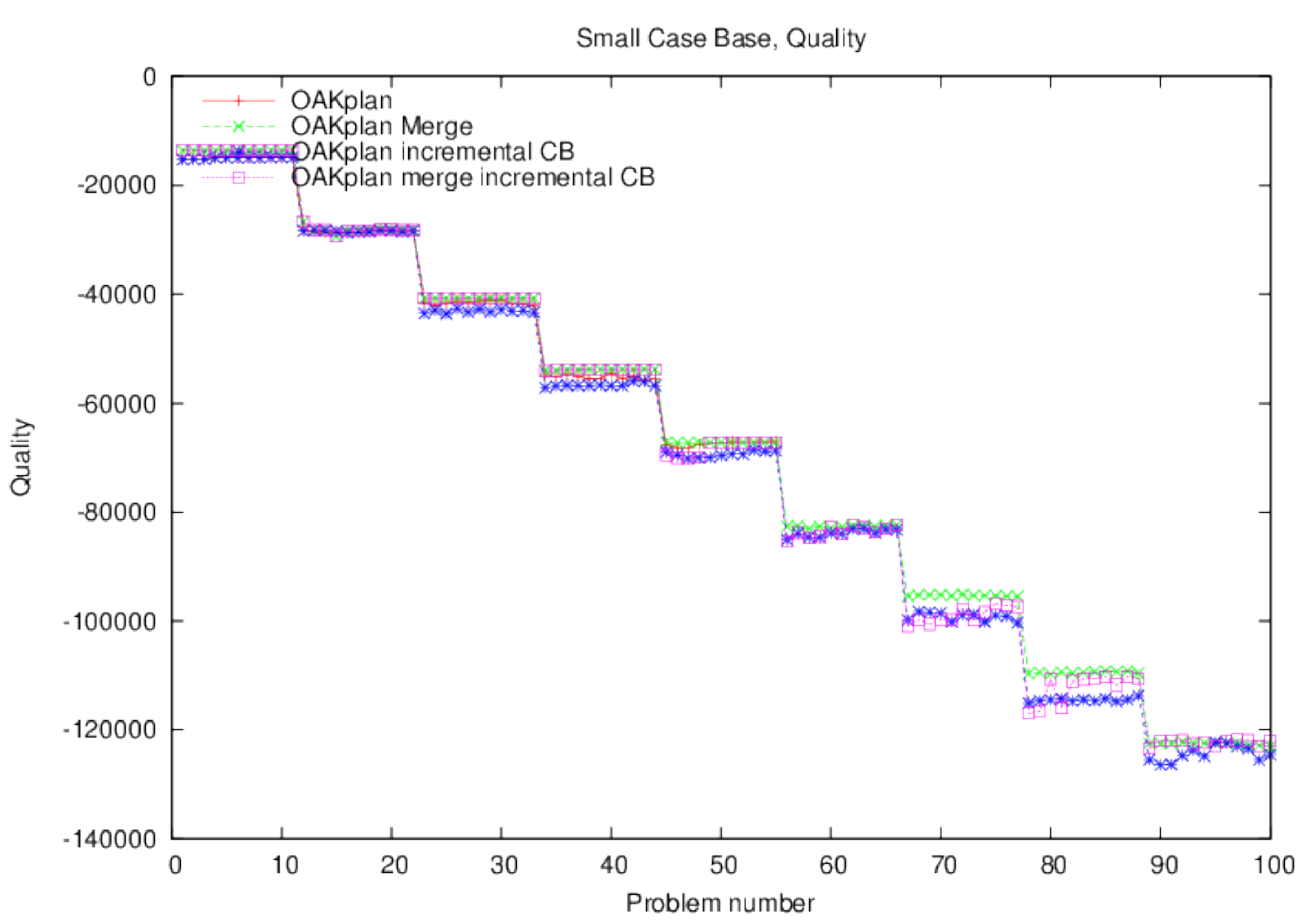}
\vspace{-0.3cm}

\\

\includegraphics[angle=-90,width=0.50\textwidth]{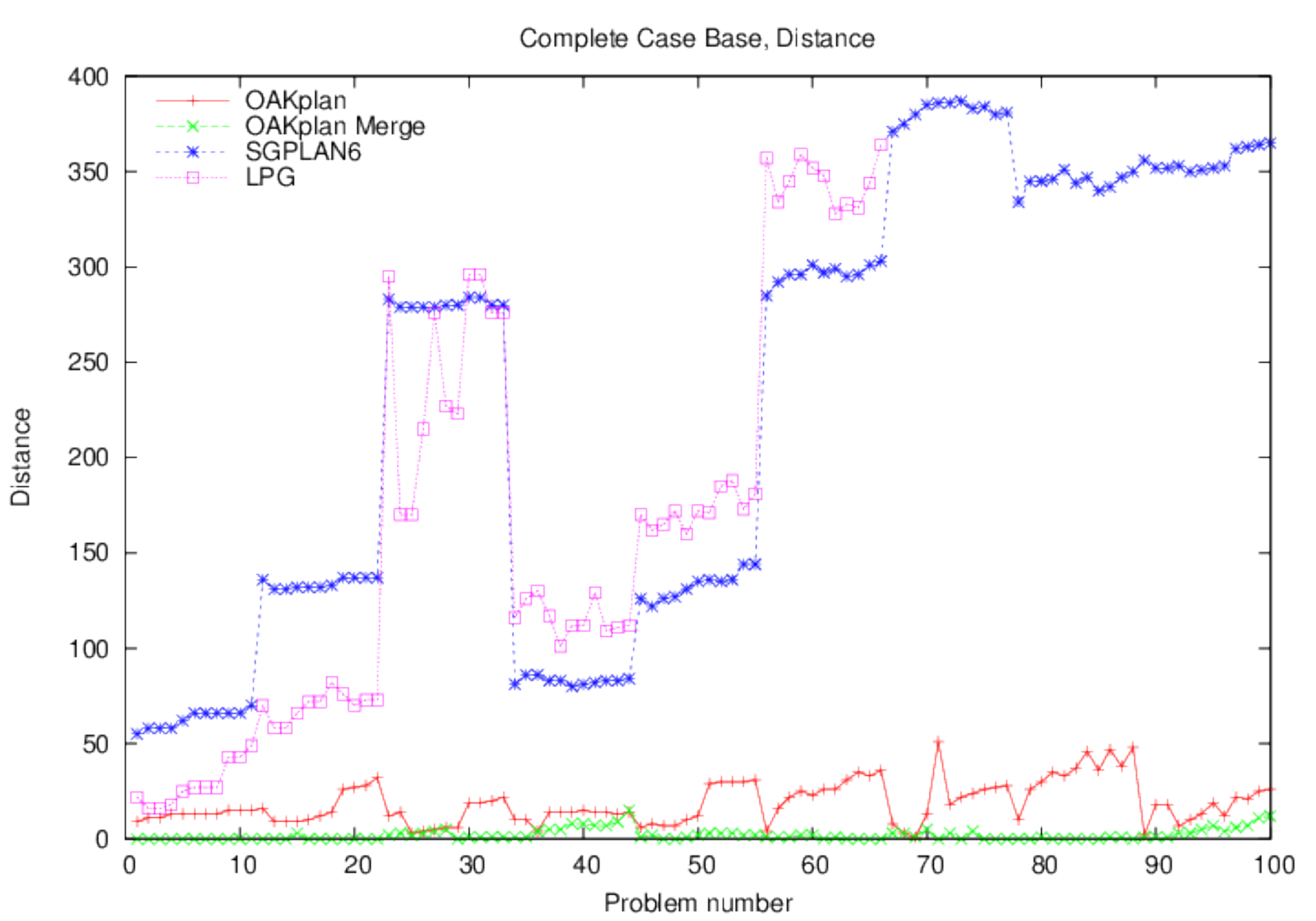}
&
\hspace{-7mm}

\includegraphics[angle=-90,width=0.50\textwidth]{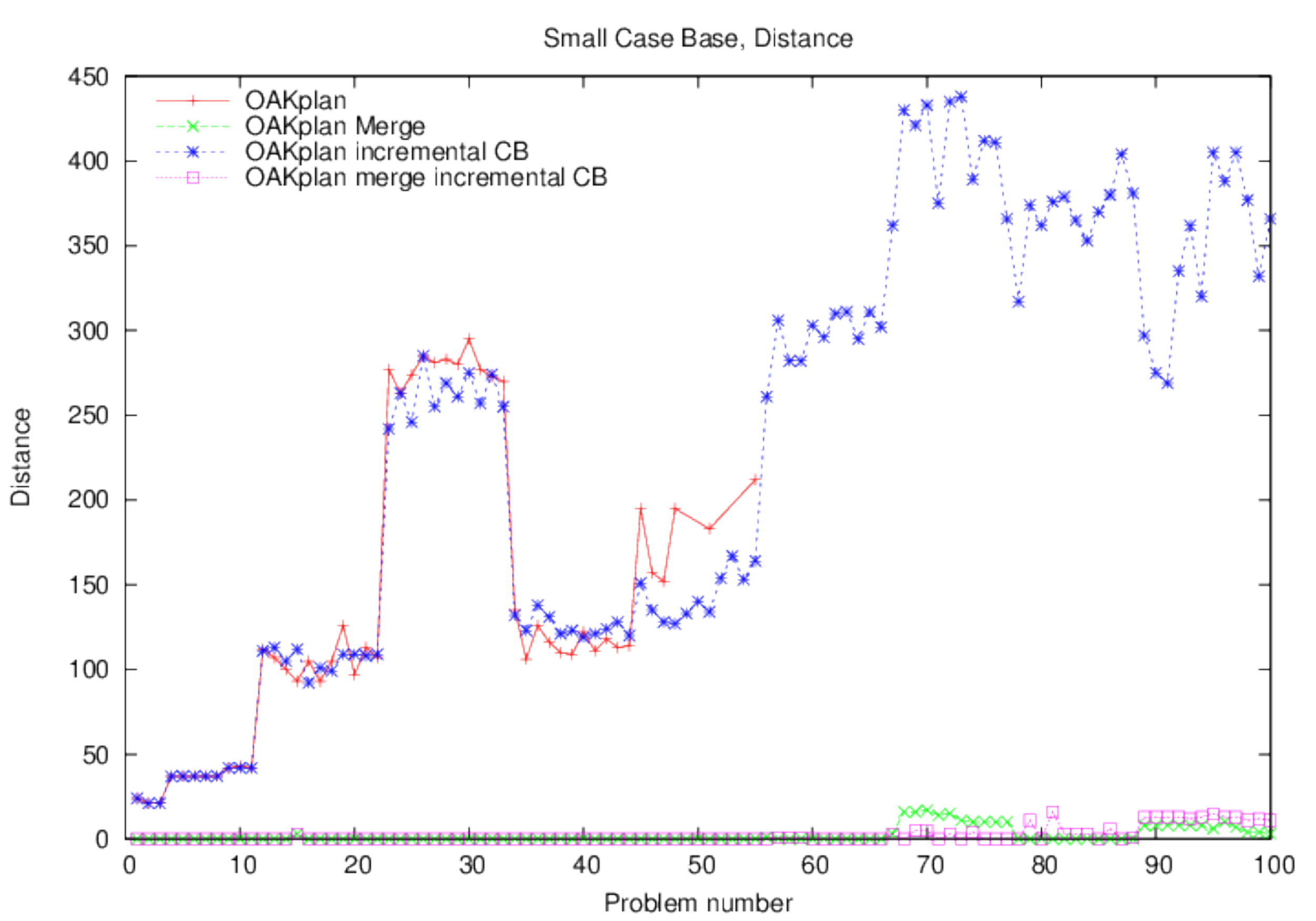}

\end{tabular}

\caption{ \label{fig:E3-all-newnames} Respectively per rows, CPU time (on a logarithmic scale), plan quality, and number of different LOs w.r.t. the plan retrieved from the library (for \OAKplan and \OAKplan-merge) or the best solution available of the base problem (for \lpg\ and \sgplan) --- small
  values are preferable except for quality. In the first column we compare the case-based approaches (\OAKplan\ and \OAKplan-merge)\emph{vs.} replanning (\lpg\ and \sgplan).
In the second column we compare the case-based approaches considering different input case bases.
}
\end{figure*}



\begin{table}[t!]  
{
\begin{center}
\scriptsize
\begin{tabular}{| l || r | r | r  | r | r | }
\hline\hline
Planner & Num. & Speed & Quality  &  Distance \\  
\hline
\OAKplan & 100 & 16.31(66.8) & -72098 ({\bf 99}) & 18.76 (12)\\
 small CB & 50 & 13.79(30.6) & -38973 (47.4) & 139 (0.63)\\
 small CB \& incr tech. &  100 & 13.5 ({\bf 94}) &  -71048 (97.3) & 233 (0.91)\\
\hline
\OAKplan-merge & 100 & 30.66 (41.7) & -70444 (95.9) & 2.06 (80.8)\\
 small CB & 100 & 40.3 (32) & -68709 (93.5) & 2.18 ({\bf 86.4})\\
 small CB \& incr tech. &  100 & 35.8 (37.6) & -69607 (94.5) & 2.21 (85.7)\\
\hline
\sgplan & 100 & 15.73 (84.8)& -69883 ({ 95.4}) & 231.5 (0.76)\\
\hline
\lpg & 66 & 67.46 (9.3) & -48315 (63.2) & 163 (0.87)\\
\hline \hline
\end{tabular}
\end{center}
}
\vspace{-3mm}
\caption{\label{table:results}
Number of problems solved, average CPU-time (seconds), average Quality and average  Distance of \OAKplan, \OAKplan-merge, {\sgplan} and \lpg.
}
\end{table}


Figure \ref{fig:E3-all-newnames} depicts the results: the time taken to
produce a solution ---the first one for \lpg, \OAKplan\ and \OAKplan-merge --- (top);
the quality of the generated routes (middle); and the stability, in
terms of distance of the new routes to the original ones (bottom). We
show the best distance and plan quality across all plans produced in
the entire optimization phase\footnote{Note that the first plan
  generated by \lpg,  \OAKplan\ and \OAKplan-merge, the best quality plan and the best
  distance plan could be different plans. It depends on the teacher's
  preferences to give more importance to the plan quality or to the
  plan stability by selecting the most appropriate solution plan
  during the validation process.}.

In the plots on the left we compare {\OAKplan} (with and without
merging techniques) \emph{vs.} \lpg\ and \sgplan\ using a ``complete''
case-base that contains all the base problems and the corresponding
solutions (the case-base for the merging variants contains also the
selected subplans of the base problems).  Here we can observe the
general good behaviour of the case-based techniques, which are
comparable in terms of CPU-time to {\sgplan} and slightly better than
the other planners in terms of plan quality for the bigger instances.
\OAKplan-merge is slightly slower than \OAKplan\ since it tries to
reuse not only the base problems but also their subplans.  The results
demonstrate that the case-based approach is at least as fast as
replanning, and sometimes faster. Obviously, plan retrieval techniques
show less useful when the changes are significant and fixing the route
requires more effort than simply discarding it and rebuilding a new
one from scratch. But the benefits for investing this effort can be
seen in terms of stability. On the other hand, the retrieval and
adaptation process sometimes comes at a price in terms of quality, as
the route is adapted to fit a new configuration rather than
constructed expressly for it. But our experiments show that the
quality for the case-based approach can be better than for replanning,
particularly in the most complex problems (see Figure
\ref{fig:E3-all-newnames}).

Finally, the best values for stability
are achieved in \OAKplan-merge.  While replanning generates routes that are
consistently very different to the original ones, the differences
between the retrieved plan and the solution plans are very small.
This high value is not particularly surprising since {\lpg} and
{\sgplan} do not know the target plans used for this comparison.
These distance values are interesting since they are a clear indicator
of the good behaviour of case-based techniques and show that the generative
approach is not feasible when we want to preserve the stability of the
plans produced.  Moreover, we can observe the extremely good behaviour
of \OAKplan-merge; its distance values are obtained considering the
number of different actions w.r.t. the matching plan provided by the
retrieval phase, which is not necessarily obtained directly by the
solution plan stored in the case base  (as in \OAKplan), but also using the different
subplans (highlighted to the teachers) obtained by the analysis of the
case-based elements that best fit the current initial state and goals.
This indicator is very appealing in an e-learning setting as
the students/teachers do not want to deal with an entirely new
learning route after a little change happens during execution. Quite
the contrary, students and teachers prefer a kind of \emph{inertia} in
the learning routes that facilitates the learning process.

In the plots on the right of Figure \ref{fig:E3-all-newnames} we analyse the behaviour of {\OAKplan} and {\OAKplan}-merge to study the impact of using a case base considering:  i) a case base created using only the smallest base problem (with 10 students),
ii) a case base where the base problems are progressively inserted
{\it after} the corresponding variants have been evaluated (it
initially contains only the smallest base problem).  In the first case,
we primarily want to evaluate the ability of the merging techniques to
reuse the solutions available in the case base at the increase of the
``differences'' (in terms of number of students) among the current
situation and the elements stored in the case base.
In particular, we want to examine the scalability in terms of students
which is extremely important in our context where a teacher could
decide to evaluate the effectiveness of an e-learning course
considering a limited number of students before using it for the whole
class.

Here we can observe the general good behaviour of \OAKplan-merge, while \OAKplan\ {\it without} merging techniques is able to solve only 50 variants with high distance values w.r.t.  the
solution plan of the base problems. In fact, \OAKplan\  can retrieve from the case base only the base problem with 10 students, whose solution plan is very different w.r.t. the
final solution plan., so we have not plotted the
distance w.r.t. the solution plan provided by the retrieval phase since
it  corresponds to the solution plan of the base problem with 10 students which is obviously extremely different w.r.t. the
final solution plan.

Regarding the tests with the {\it incremental} case base, we want to analyse
the behaviour of {\OAKplan} considering a case base that contains
elements which are structurally not to much different w.r.t.  the
current situation. For example, considering the solution of the
variants with 50 students, the case base does not contain the base
problem with 50 students but it contains the base problems with
10, 20, 30 and 40 students.
Here we want to examine the situation where a teacher has already used
a course in different classes and wants to reuse the stored
experiences in a new (slightly bigger) class.
As expected, the behaviour of
{\OAKplan}-merge does not change significantly neither in term of
CPU-time nor in term of distance and plan quality; on the contrary,
the CPU-time of {\OAKplan} {\it without} merging techniques decreases
significantly since it can replan starting from a case base element
with a slightly lower number of students w.r.t. the current situation;
moreover it is now able to
solve all the variants considered.  Regarding the plan differences
w.r.t. the solution of the base problems we can observe values which
are similar to the ones obtained for replanning. This is not
surprising since the elements stored in the case base are different
w.r.t. the current situation and {\OAKplan} does not know the target
solution plan; on the contrary the performances of {\OAKplan}-merge
are extremely good, both considering the case base with only the
smallest base problems and the incremental case base. However, it is
important to point out that in this case the comparison in terms of
plan distances are performed considering directly the plan provided by
the retrieval phase. It is up to the teacher to decide if (s)he wants
to validate elements that considers only previously executed courses or
also subparts of them.

Globally we can observe that the use of plan merging techniques are
potentially very effective in our execution context, allowing to obtain
efficiently new e-learning routes which are very similar to the
combination of previously executed ones (or subpart of them).  This is
extremely important since it allows to highlight to the teachers the
changes w.r.t. already executed learning routes, facilitating the
validation process. Moreover, the stored plans can also contain some
notes, regarding for example the pedagogical motivations associated to
the selection or combination of specific LOs, annotated by the teacher during the creation of the original learning route or during previously executions of the learning route; in this case, these notes can be easily reexamined
by the teachers facilitating the learning route validation process.

\section{Related Work}
\label{sec:RelatedWork}

In the following section we examine the most relevant case-based
planners considering their retrieval, adaptation and storage
capabilities. Moreover, we present an empirical comparison of the
performance of  {\OAKplan} vs. the {\sc far-off} system and some
comments on the advantages of  {\OAKplan} with respect to other case-based planners.

Some CBP systems designed in the past do not consider any generative planning
in their structure, and find a solution only by the cases stored in the
case base. These CBP systems are called {\it reuse-only} systems.  As
reuse-only systems cannot find any planning solution from scratch, they cannot find a solution unless  they  find a proper case in
the case base that can be adapted through single rules.  An alternative
approach to reuse-only systems is the {\it reuse-optional} approach, which
uses a generative planning system that is responsible to adapt the retrieved
cases. This feature allows a CBP system to solve problems that cannot be
solved only by using stored cases and simple rules in the adaptation phase.
Empirically, a great number of reuse-optional CBP systems has shown
that the use of a case base can permit them to perform better in
processing time and in a number of planning solutions than the
generative planning that they incorporate.

Obviously the retrieval phase critically affects the systems
performance; it must search in a space of cases in order to choose a
good one that will allow the system to solve a new problem
easily. In order to improve efficiency in the retrieval phase, it is
necessary either to reduce the search space or to design an accurate
similarity metric.  Reducing the search space, only a suitable subset
of cases will be available for the search process and an accurate
similarity metric will choose the most similar case to decrease the
adaptation phase effort.
In the literature there are different domain dependent and a few domain
independent plan adaptation and case-based planning systems, which mostly
use a search engine based on a space of states
~\cite{GerSer-aips00,HanWel95,TonidandelR02,KrogtW-icaps05}.  An
alternative approach to planning with states is that of  plan-space
planning or hierarchical systems ~\cite{bergmann95building} that
search in a space of plans and have no goals, but only tasks to be
achieved.  Since tasks are semantically different from goals, the
similarity metric designed for these CBP systems is also different
from the similarity rules designed for state-space based CBP systems.
For a detailed analysis of case-based and plan adaptation techniques
see the papers of Spalazzi \cite{Spalazzi2001} and Munoz-Avila \& Cox
\cite{munoz-avila-cox-2007}.

\ifTR
The {\sc chef} system \cite{Ham90} is the first application of CBR in
planning and it is a reuse-only system which is important especially from a
historical point of view. It solves problems in the domain of Szechwan cooking
and is equipped with a set of plan repair rules that describe how a
specific failure can be repaired.  Given a goal to produce a dish with
particular properties, {\sc chef} first tries to anticipate any problems or
conflicts that may arise from the new goal and repairs problems that did not
arise in the baseline scenario. It then executes the plan and, if execution
results in a failure, a repair algorithm analyses the failure and builds an
explanation of the reason why the failure has occurred. This explanation
includes a description of the steps and states leading towards the failure as
well as the goals that these steps tried to realise. Based on the explanation,
a set of plan repair strategies is selected and instantiated to the specific
situation of the failure. After choosing the best of these instantiated repair
strategies, {\sc chef} implements it and uses the result of the
failure analysis to improve the index of this solution so that it will not be
retrieved in situations where it will fail again.

Much attention has been given to research that designs suitable similarity
metrics. It focuses on choosing the most adaptable case as the most similar
one, such as the {\sc dial} \cite{leake97casebased} and {\sc D\'{e}j\`{a}Vu}
\cite{smyth98adaptationguided} systems. The {\sc dial} system is a case-based
planner that works in disaster domains where cases are schema-based
episodes and uses a similarity assessment approach, called {\it RCR}, which
considers an adaptability estimate to choose cases in the retrieval phase.
Our similarity functions differ from the RCR method since they are based on a
domain knowledge that is available in action definitions, while the RCR method
uses the experience learned from the adaptation of previous  utilisation of
cases. They also differ in their applicability because the RCR method
considers specifically disaster domains while our approach is suitable for
domain independent planning.

Similarly, the {\sc D\'{e}j\`{a}Vu} system operates in design domains and uses
an {\it adaptation-guided retrieval} (AGR) procedure to choose cases that are
easier to be adapted.  The AGR approach in the {\sc D\'{e}j\`{a}Vu} system uses
additional domain knowledge, called capability knowledge, which is similar to
that used to solve conflicts in partial-order planning systems. This
additional knowledge allows to identify the type and the functionality of a
set of transformations, which are performed by actions, through a collection
of agents called specialists and strategies.  It must be well specified
so as to maximise the AGR performance. Our similarity functions differ
from the AGR approach because we do not use any domain knowledge besides
that obtained from actions and states, which is the minimal knowledge
required to define a domain for planning systems.

The {\sc plexus} system \cite{Alterman90} confronts with the problem
of ``adaptive planning'', but also addresses the problem of runtime
adaptation to plan failure.  {\sc plexus} approaches plan adaptation
with a combination of tactical control and situation matching. When
plan failure is detected it is classified as either beginning a
failing precondition, a failing outcome, a case of differing goals or
an out-of-order step. If we ignore how to manage  incomplete
knowledge, the repair strategy involves the fact of replacing a failed
plan step with one that might achieve the same purpose. It uses a
semantic network to represent abstraction classes of actions that
achieve the same purpose.

The {\sc gordius} \cite{Simmons88} system is a transformational planner that
combines small plan fragments for different (hopefully independent)
aspects of the current problem. It does not perform an anticipation
analysis on the plan, on the contrary it accepts the fact that the
retrieved plan will be flawed and counts on its repair heuristics to
patch it; in fact, much of the {\sc gordius} work is devoted to developing a
set of repair operators for quantified and metric variables.
The previous approaches differ with respect to {\sc  OAKplan} fundamentally
because they are domain dependent planners; on the contrary  {\sc  OAKplan}
uses only the domain and planning problems descriptions.

\fi

Three interesting works developed at the same time adopt similar assumptions:
the {\sc priar} system \cite{KamHen92}, the {\sc spa} system \cite{HanWel95}
and the Prodigy/Analogy system ~\cite{veloso92learning,Veloso:1994:PLA}. {\sc
  priar} uses a variant of Nonlin \cite{Tate-ijcai77}, a hierarchical planner,
whereas {\sc spa} uses a constraint posting technique similar to Chapman's
Tweak~\cite{Chapman87} as modified by McAllester and
Rosenblitt~\cite{McAllester91}. {\sc priar}'s plan representation and thus its
algorithms are more complicated than those of {\sc spa}. There are three
different types of validations (filter condition, precondition, and phantom
goal) as well as different {\em reduction levels} for the plan that represents a
hierarchical decomposition of its structure, along with five different
strategies for repairing validation failures. In contrast to this
representation the plan representation of {\sc spa} consists of causal links
and step order constraints.  The main idea behind the {\sc spa} system that
separates it from the systems mentioned above is that the process of plan
adaptation is a fairly simple extension of the process of plan generation. In
the {\sc spa} view, plan generation is just a special case of plan adaptation
(one in which there is no retrieved structure to exploit).  With respect to
our approach that defines a matching function $\mu$ from $\Pi$ to $\Pi'$ that
maximises the similarity function $simil_\mu$, it should be noted that in {\sc
  priar} and {\sc spa} the conditions for the initial state match are slightly
more complicated. In {\sc priar} the number of {\em inconsistencies in the
validation structure} of the plan library is minimised; in {\sc spa} the
number of violations of preconditions in the plan library is
maximised. Moreover the problem-independent matching strategy implemented in
{\sc spa} runs in exponential time in the number of objects since it simply
evaluates all possible mappings.  On the contrary we compute an approximate
matching function in polynomial time and use an accurate plan evaluation
function on a subset of the plans in the library.

The Prodigy/Analogy system also uses a search oriented approach to
planning. A library plan (case) in a transformational or
case-based planning framework stores a solution to a prior problem
along with a summary of the new problems for which it would be a suitable
solution, but it contains little information on the process
that generates the solution. On the other hand derivational analogy
stores substantial descriptions of the adaptation process decisions in the
solution, whereas Veloso's system records more information at
each choice point than  {\sc spa} does, like a list of failed alternatives.
An interesting similarity rule in the plan-space approach is presented
in the {\sc caplan/cbc} system \cite{Munoz-Avila96} which extends the
similarity rule introduced by the Prodigy/Analogy system
~\cite{veloso92learning,Veloso:1994:PLA} by using feature weights in
order to reduce the errors in the retrieval phase.
\ifTR
These feature weights are learned and recomputed according to
the performance of the previous retrieved cases and we can note that
this approach is similar to the RCR method used by the {\sc dial}
system in disaster domains.  \fi
There are two important differences between our approach and the
similarity rules of  {\sc caplan/cbc}, one of which
is that the former is designed for state-space planning and the latter
for plan-space planning. Another difference is that our retrieval
function does not need to learn any knowledge to present an accurate
estimate: our retrieval method only needs the knowledge that can be
extracted from the problem description and the actions of the planning
cases.

\ifTR
O-Plan \cite{CurrieT91,drabble97repairing} is based on the strategy of
using plan repair rules as well. The effects of every action are
confirmed while execution is performed. A repair plan formed by
additional actions is added to the plan every time a failing effect is
necessary in order to execute some other actions. We call repair plans
the prebuilt ones which are in a position to repair a series of
failure conditions. For instance, we can have repair plans including a
plan to replace either a flat tyre or a broken engine.  When an
erroneous condition is met, the plan is no longer executed but a
repair plan is inserted and executed. When the repair plan is
complete, the regular plan is executed once more. Failures are
repaired by O-Plan by adding actions. It follows that it does not use
either unrefinements or requires a history. However it is not complete
and there are some failures which cannot be repaired.
\fi

{\sc mlr} \cite{Nebel95} is another case-based system and it is based
on a proof system.  While retrieving a plan from the library that has
to be adapted to the current world state, it makes an effort to employ
the retrieval plan as if it were a {\em proof} to set the goal
conditions from the start. Should this happen, there is no need for
any iteration to use the plan, otherwise, the outcome is a failed
proof that can provide refitting information. On the basis of the
failed proof, a plan skeleton is built through a {\it modification
  strategy} and it makes use of the failed proof to obtain the parts
of the plan that are useful and removes the useless parts. After the
computation of this skeleton, gaps are filled through a refinement
strategy which makes use of the proof system. Although our object
matching function is inspired to the Nebel \& Koehler's formalisation,
our approach significantly differs from theirs since they do not
present an effective domain independent matching function. In fact,
their experiments exhibit an exponential run time behaviour for the
matching algorithm they use, instead we show that the retrieval and
matching processes can be performed efficiently also for huge plan
libraries.  The matching function formalisation proposed by Nebel \&
Koehler also tries to maximise first the cardinality of the common
goal facts set and second the cardinality of the common initial facts
set. On the contrary we try to identify the matching function $\mu$
that maximise the $simil_\mu$ similarity value which considers both
the initial and goal relevant facts and an accurate evaluation
function based on a simulated execution of the candidate plans is used
to select the best plan that has to be adapted.

 Nebel \& Koehler \cite{Nebel95} present an interesting comparison of
 the {\sc mlr}, {\sc spa} and {\sc priar} performance in the
 BlocksWorld domain considering planning instances with up to {\em $8$
   blocks}.  They show that also for these small sized instances and
 using a {\em single} reuse candidate the matching costs are already
 greater than adaptation costs. When the modification tasks
 become more difficult, since the reuse candidate and the new planning
 instance are structurally less similar, the savings of plan
 modification become less predictable and the matching and adaptation
 effort is higher than the generation from scratch.  On the contrary
 {\OAKplan} shows good performance with respect to plan generation
 and our tests in the BlocksWorld domain consider instances
 with up to {\em $140$ blocks} and a plan library with  {\em ten thousands}
  cases.

\ifTR
The LPA* algorithm is used by the {\sc Sherpa} replanner
\cite{LikhachevK05}.  This algorithm was originally bound to repair path
plan and backtrack to a partial plan having the same heuristic value
as before the unexpected changes did in the world using the unrefinement
step once. {\sc Sherpa} is not useful to solve every repair problem,
owing to the unrefinement strategy and the single application
thereof. Its use is restricted to those problems whose actions are no
longer present in the domain description. It follows that through the
unrefinement step unavailaible actions are removed.

   The Replan \cite{BoellaD02} model of plans is similar to the
plans used in the hierarchical task network (HTN) formalism
\cite{erol.hendl-1994}. A task network is a description of a possible
way to fulfil a task by doing some subtasks, or, eventually
(primitive) actions. For each task at least one of such task networks
exists. A plan is created by choosing the right task networks for each
(abstract) task chosen, until each network consists of only
(primitive) actions. Throughout this planning process, Replan
constructs a derivation tree that includes all tasks chosen, and shows
how a plan is derived.  Plan repair within Replan is called
partialisation. For each invalidated leaf node of the derivation
tree, the (smallest) subtree that contains this node is removed.
Initially, such an invalid leave node is a primitive action and the
root of the corresponding subtree is the task containing this action.
Subsequently a new refinement is generated for this task.  If the
refinement fails, a new round is started in which task subtrees
that are higher in the hierarchy are removed and regenerated. In the
worst case, this process continues until the whole derivation tree is
discarded.

\begin{figure}[t!]
\begin{center}
 \includegraphics[angle=-90,width=0.7\textwidth]{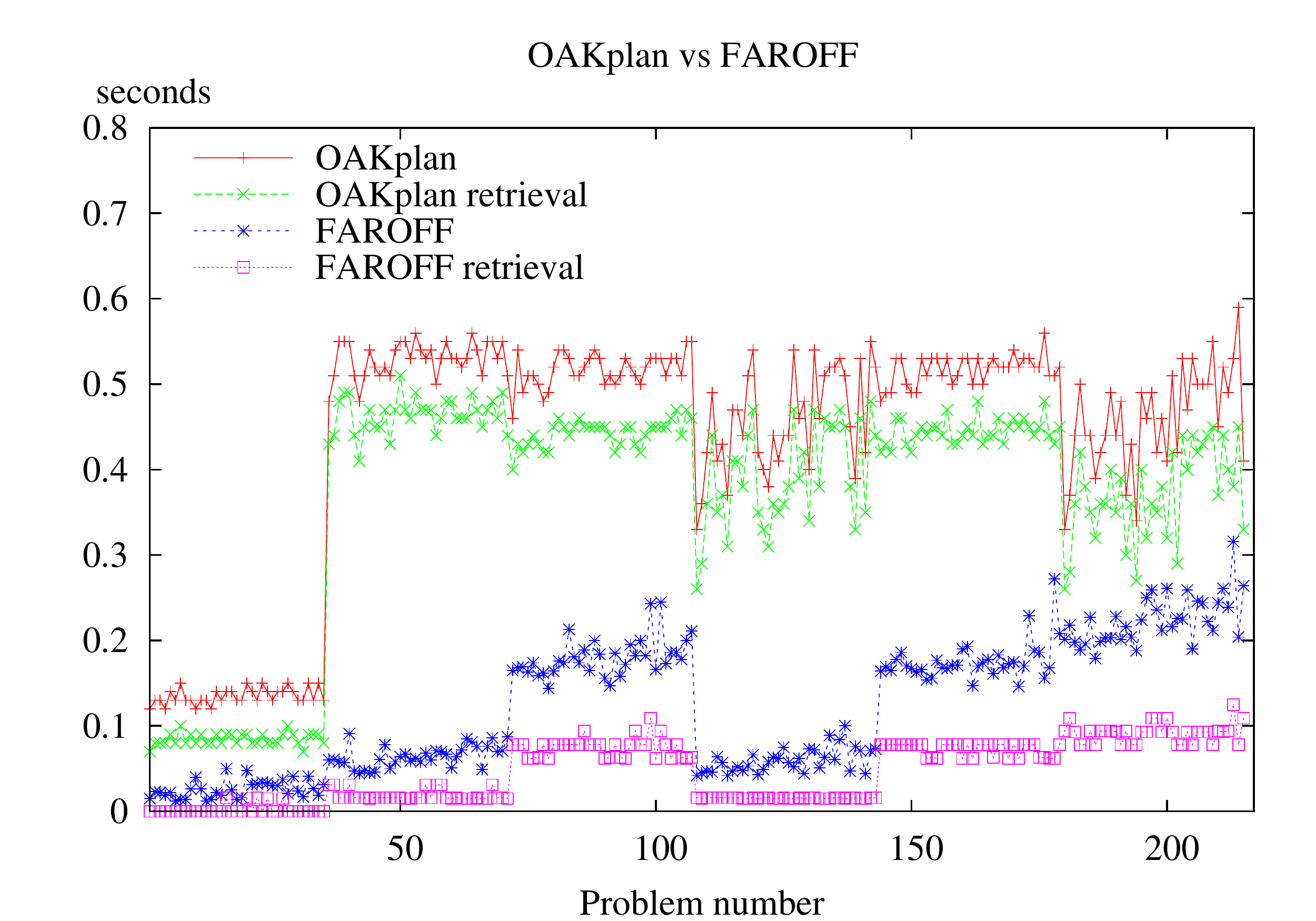}
\\
\vspace{-3mm}
\includegraphics[angle=-90,width=0.7\textwidth]{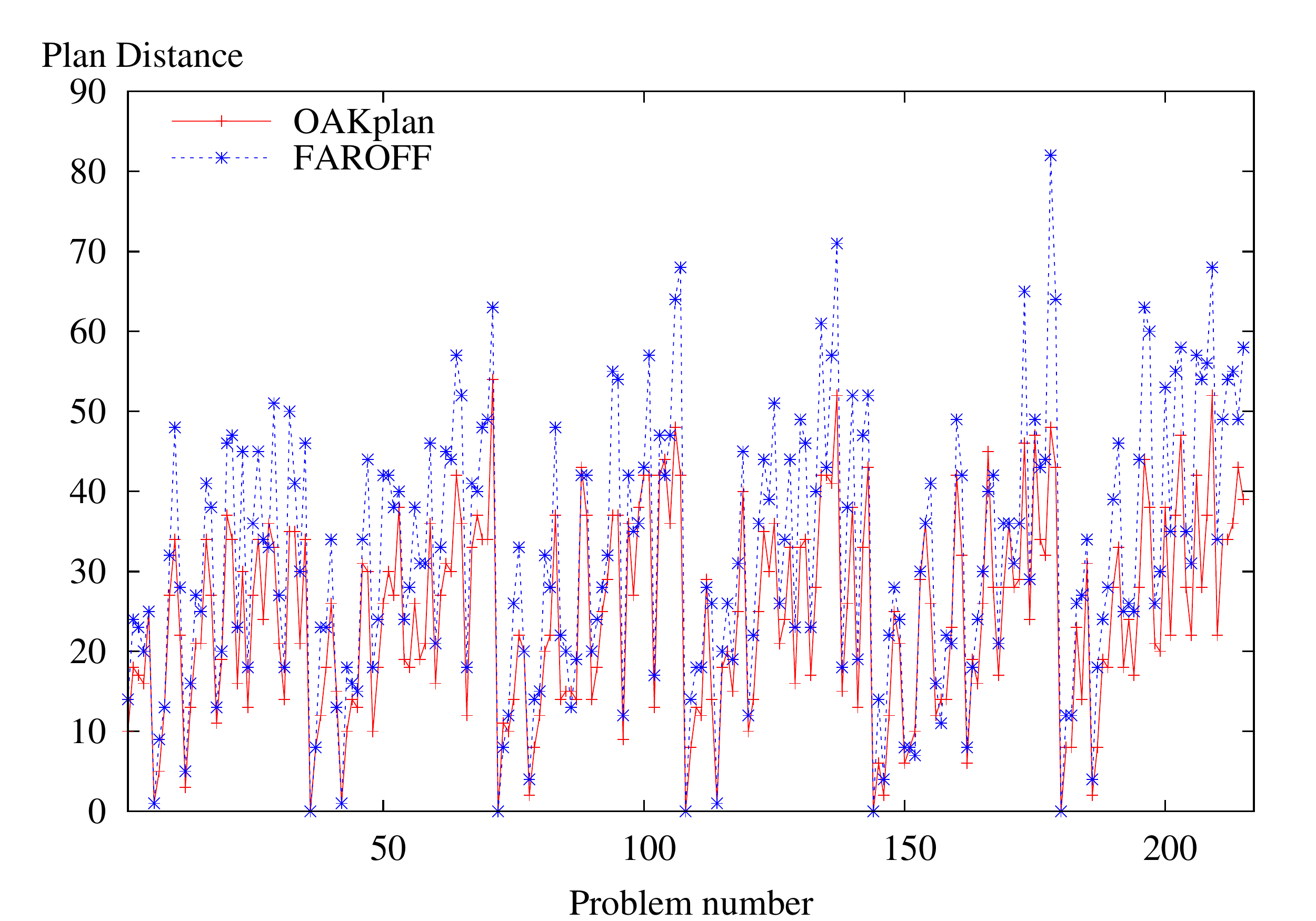}
\\
\vspace{-3mm}
\includegraphics[angle=-90,width=0.7\textwidth]{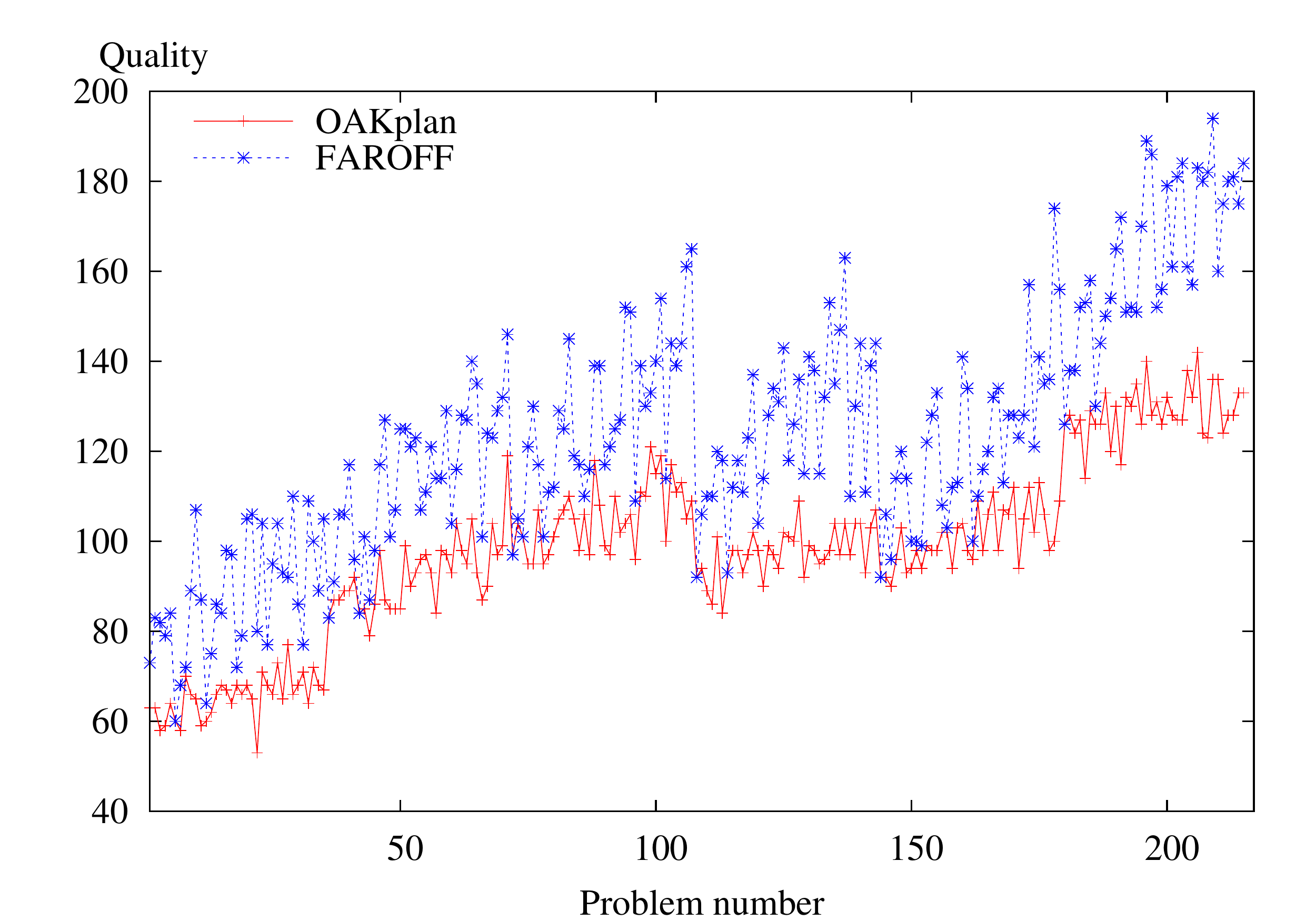}
\end{center} %
\vspace{-4mm}
\caption{\ifTR \small \fi\label{fig:OAKplan-vs-FAROFF}
CPU-time, number of different actions and plan qualities for the
Logistics variants.  Here we examine {\sc OAKplan} vs. {\sc far-off}.
}

\end{figure}

\fi

A very interesting case-based planner is the {\sc
  far-off}\footnote{{\sc far-off} is available at
  http://www.fei.edu.br/$\sim$flaviot/faroff.} ({\it Fast and Accurate
  Retrieval on Fast Forward}) system \cite{TonidandelR02}.  It uses a
generative planning system based on the FF planner \cite{HofNeb-JAIR01}
to adapt similar cases and a similarity metric, called ADG (Action
Distance-Guided), which, like {\sc EvaluatePlan}, determines
the adaptation effort by estimating the number of actions
that is necessary to transform a case into a solution of the
problem. The ADG similarity metric calculates two estimate values of
the distance between states.  The first value, called {\it initial
  similarity value}, estimates the distance between the current
initial state $I$ and the initial state of the case $I_\pi$ building a
relaxed plan having $I$ as initial state and $I_\pi$ as goal state.
Similarly the second value, called {\it goal similarity value},
estimates the distance between the final state of the case and the
goals of the current planning problem.
Our {\sc EvaluatePlan} procedure evaluates instead
every single inconsistency that a case base solution plan determines in the
current world state $I$.

The {\sc far-off} system uses a new competence-based method, called
{\it Footprint-based Retrieval} \cite{SmythM99a}, to reduce the space of cases
that will be evaluated by ADG. The Footprint-based Retrieval is a
competence-based method for determining groups of footprint cases that
represent a smaller case base with the same competence of the original
one. Each footprint case has a set of similar cases called {\it Related Set}
\cite{SmythM99a}. The union of footprint cases and Related Set is the original
case base.  On the contrary \OAKplan~uses a much more simple procedure based
on the $simil^{ds}$ function to filter out irrelevant cases. The use of
Footprint-based Retrieval techniques and case base maintenance policies in
\OAKplan~is left for future work.
It is important to point out that the retrieval phase of {\sc
  far-off} does not use any kind of abstraction to match cases and problems.

The {\sc far-off} system retrieves the most similar case, or the
ordered $k$ most similar cases, and shifts to the adaptation phase.
Its adaptation process does not modify the retrieved case, but only
completes it; it will only find a plan that begins from the current
initial state and then goes to the initial state of the case, and
another plan that begins from the state obtained by applying all the
actions of the case and goes to a state that satisfies the current
goals $G$. Obviously, the completing of cases leads the {\sc far-off}
system to find longer solution plans than generative planners, but it
avoids wasting time in manipulating case actions in order to find
shorter solutions length.  To complete cases, the {\sc far-off} system
uses a FF-based generative planning system, where the solution is
obtained by merging both plans that are found by the FF-based
generative planning and the solution plan of the planning case
selected. On the contrary \OAKplan~uses the \lpg-adapt adaptation
system, which uses a local search approach and works on the whole
input plan so as to adapt and find a solution to the current planning
problem.

In Figure \ref{fig:OAKplan-vs-FAROFF} we can observe the behaviour of
\OAKplan~vs {\sc far-off} considering different variants of the
greater case bases provided with the {\sc far-off} system in the
Logistics domain;\footnote{We have used the case bases for the
  logistics-16-0, logistics-17-0 and logistics-18-0 Logistics IPC2
  problems. For each problem considered the {\sc far-off} system must
  have a case base with the same structure to perform tests. More than
  700 cases belong to each case base and for each case base we have
  selected two planning cases and randomly generated $36$ variants.  }
similar results have been obtained in the BlocksWorld, DriverLog and
ZenoTravel domains.
Globally, we can observe that {\sc far-off} is always faster than
\OAKplan~both considering the retrieval and the total adaptation time
although also the \OAKplan~CPU-time is always lower than $0.6$
seconds.  Considering \OAKplan, most of the CPU-time is devoted to the
computation of the matching functions which are not computed by {\sc
  far-off} since it simply considers the {\it identity} matching
function that directly assigns the objects of the case base to those
of the current planning problem with the same name. In fact, it
does not consider objects which are not already present in the case
base and, to overcome this limitation, the variants used in this test
are directly obtained by the problems stored in the case bases.

Regarding the plan qualities\footnote{In STRIPS domains the plan
  quality is obtained by considering the number of actions in the
  solution plan.} and the plan distances, it is important to point out
that for each variant solved by \OAKplan~we consider only the first
solution produced since {\sc far-off} does not perform a plan
optimisation process. However {\OAKplan} is able to obtain better plans both
considering the plan quality and the plan distance values. Globally,
\OAKplan~is able to find plans with $20\%$ better quality and $24\%$
better plan distances. Moreover further improvements on plan qualities
and distance values of \OAKplan~could be obtained by performing the
optimisation process of \lpg-adapt.

Finally, note that in this experiment we have
used the case bases provided by {\sc far-off} which contain $700$
elements each and the corresponding cases are generated by creating randomly
planning problems all with the same configuration: same objects,
trucks and airplanes simply disposed in different ways.  This kind of
experiment is highly unfavourable to {\OAKplan} since our first
screening procedure cannot filter out a significant number of cases as
they all have the same structure. On the contrary, in the experiments
described in the previous sections the case bases used by {\OAKplan}
in the standard configuration (not the ``small'' versions) are not
constrained to a particular planning problem but they have been
generated by considering all the different planning problems
configurations used in the International Planning Competitions. This
is a much more realistic situation, where the cases are added to the
case base when the planning problems provided by the users are
resolved as time goes by.

\newpage



{\small
\bibliography{./biblio6}
\bibliographystyle{plain}
}

\end{document}